\documentclass[journal]{IEEEtran}
\usepackage{amsmath,amsfonts}
\usepackage{algorithmic}
\usepackage{algorithm}
\usepackage{array}
\usepackage{textcomp}
\usepackage{stfloats}
\usepackage{url}
\usepackage[citecolor=green]{hyperref}
\usepackage{verbatim}
\usepackage{graphicx}
\usepackage[sort]{cite}
\usepackage{subfigure}
\usepackage{multirow}
\usepackage{epsfig}
\usepackage{amsmath}
\usepackage{amssymb}
\usepackage{wrapfig}
\hypersetup{colorlinks=true,linkcolor=black}
\begin{document}

\title{Multi-scale Context-aware Network with Transformer for Gait Recognition}

\author{Duowang~Zhu, Xiaohu~Huang, Xinggang~Wang, Bo~Yang, Botao~He, Wenyu~Liu, and Bin~Feng
\thanks{Duowang~Zhu, Xiaohu~Huang, Xinggang~Wang, Wenyu~Liu and Bin~Feng are with the Hubei Key Laboratory of Smart Internet Technology, and also with the School of Electronic Information and Communications, Huazhong University of Science and Technology, Wuhan 430074, China.
}
\thanks{Bo~Yang is with Wuhan FiberHome Digital Technology Co., Ltd. And Botao He is with CICT Mobile Communication Technology Company Ltd.}
\thanks{Duowang Zhu and Xiaohu Huang contributed equally to this work.}
\thanks{Bin Feng (\protect{fengbin@hust.edu.cn}) is the corresponding author.}}

\markboth{IEEE TRANSACTIONS ON MULTIMEDIA,~Vol.~XX, No.~XX, XX~2023}%
{Zhu \MakeLowercase{\textit{et al.}}: Multi-scale Context-aware Network with Transformer for Gait Recognition}

% \IEEEpubid{0000--0000/00\$00.00~\copyright~2023 IEEE}

\maketitle

\begin{abstract}
Although gait recognition has drawn increasing research attention recently, since the silhouette differences are quite subtle in spatial domain, temporal feature representation is crucial for gait recognition. Inspired by the observation that humans can distinguish gaits of different subjects by adaptively focusing on clips of varying time scales, we propose a multi-scale context-aware network with transformer (MCAT) for gait recognition. MCAT generates temporal features across three scales, and adaptively aggregates them using contextual information from both local and global perspectives. Specifically, MCAT contains an adaptive temporal aggregation (ATA) module that performs local relation modeling followed by global relation modeling to fuse the multi-scale features. Besides, in order to remedy the spatial feature corruption resulting from temporal operations, MCAT incorporates a salient spatial feature learning (SSFL) module to select groups of discriminative spatial features. Extensive experiments conducted on three datasets demonstrate the state-of-the-art performance. Concretely, we achieve rank-1 accuracies of 98.7\%, 96.2\% and 88.7\% under normal-walking, bag-carrying and coat-wearing conditions on CASIA-B, 97.5\% on OU-MVLP and 50.6\% on GREW. The source code will be available at \url{https://github.com/zhuduowang/MCAT.git}.
\end{abstract}

\begin{IEEEkeywords}
Gait Recognition, Temporal Relation Modeling, Spatial Feature Preserving.
\end{IEEEkeywords}

\section{Introduction}
\label{sec:introduction}
\IEEEPARstart{G}{ait} recognition is a biometric technology that identifies individuals based on their walking patterns. It has shown great potential in applications related to public security \cite{larsen2008gait, bouchrika2011using, yang2014influence, yang2014variability} and  identity recognition \cite{macoveciuc2019forensic, balazia2017human, premalatha2020improved}. Despite the increasing research attention on gait recognition, learning temporal feature representation is crucial due to the subtle differences in the spatial domain.

\begin{figure}[ht]
\label{motivation}
\centering
\subfigure[Two sequences from subject '53' and '119' on CASIA-B can be distinguished relying on short-term temporal clues, e.g., several frames at the beginning.]{
\label{motivation_a}
\begin{minipage}[h]{\linewidth}
\centering
\includegraphics[width=0.9\linewidth]{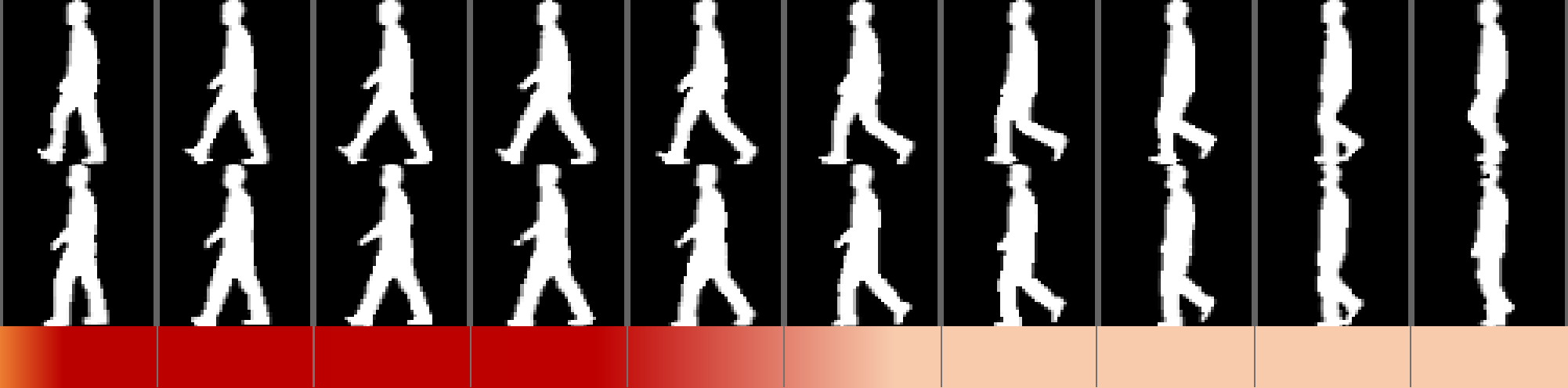}
%\caption{fig1}
\end{minipage}%
}

\subfigure[Two sequences from subject '39' and '77' on CASIA-B, which have to be distinguished relying on long-term temporal clues, e.g., all of the frames.]{
\begin{minipage}[h]{\linewidth}
\label{motivation_b}
\centering
\includegraphics[width=0.9\linewidth]{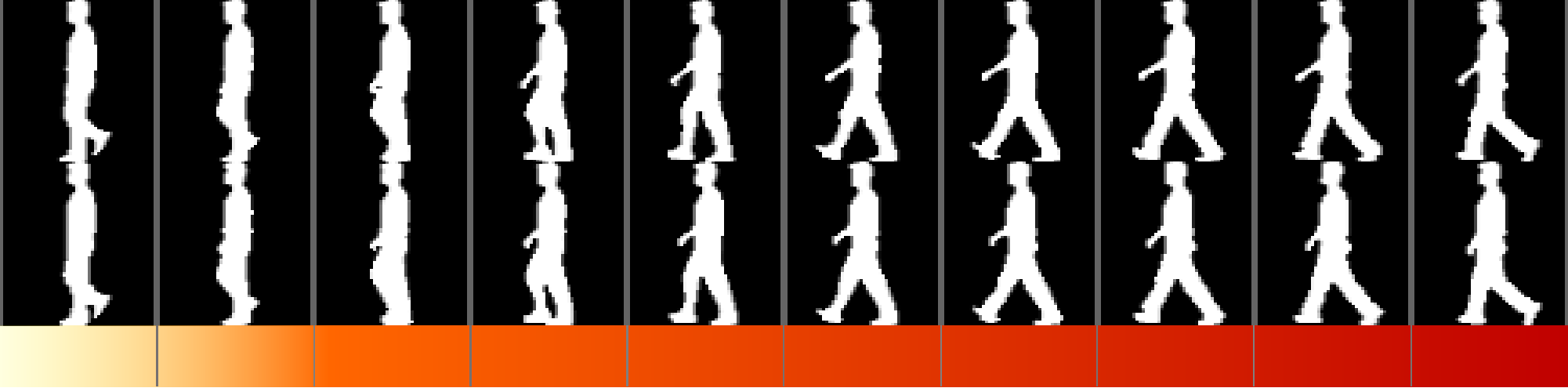}
%\caption{fig1}
\end{minipage}%
}%
\centering
\caption{Illustration that humans are capable of distinguishing gaits of different subjects by adaptively focusing on temporal fragments with different time scales. Color bar indicates the distribution of human attention. Darker color represents a higher level of attention paid to corresponding frames. Best viewed in color.}
\label{fig:motivation}
\end{figure}

Furthermore, as mentioned in \cite{fan2020gaitpart}, different body parts exhibit varied motion patterns, thus require temporal modeling to take multi-scale representation into consideration. Currently, multi-layer convolutions are commonly utilized in current methods to model multi-scale temporal information. These methods aggregate multi-scale features through summation \cite{fan2020gaitpart, wu2020condition, zhang2019cross, zhang2020learning}, or concatenation \cite{lin2020gait, GaitGL}. However, these fixed aggregation methods lack the flexibility to adapt to variations of complex motion and realistic factors, i.e., self occlusion between body parts and changes in camera viewpoints. Consequently, this limitation hinders the performance, particularly considering that gait is a kind of fine-grained motion pattern, and subject identification relies on the varied expression of customized motion on specific body parts.

It can be seen from life experience that human is able to distinguish gait sequences of different subjects by selectively attending to temporal fragments characterized by distinct time scales. A qualitative illustration is given in Figure. \ref{fig:motivation}, where voting results from 15 volunteers are used to calculate the focus distribution. In Figure. \ref{motivation_a}, the differences between the two gait sequences are so obvious that we can distinguish them by observing fewer frames from the beginning. On the contrary, in Figure. \ref{motivation_b}, differences between two sequences are quite subtle that we have to observe more frames to distinguish them. Therefore, in this situation, relying solely on short-term clues is inadequate for distinguishing between the two subjects, but long-term features need to be considered since they provide richer temporal information. Consequently, the adaptation of multi-scale temporal features facilitates a flexible focus along the temporal dimension, thereby providing a novel perspective for gait modeling.
 
 Motivated by such observations, we propose a Multi-scale Context-aware Network with Transformer (\textbf{MCAT}) for gait recognition. The core idea of this method is to integrate multi-scale temporal features by considering the contextual information along temporal dimension, which allows information exchange among different scales from both local and global perspectives. Here, contextual information is obtained by evaluating the local and global relations among multi-scale temporal features, which reflects diverse motion information existing in context semantics. MCAT produces temporal features for each frame in three scales, i.e., frame-level, short-term and long-term, which are complementary to each other. The frame-level features retain frame characteristics at each time instant. The short-term features capture local temporal contextual clues, which are sensitive to temporal locations and contribute to model micro motion patterns. The long-term features, representing motion characteristics across all frames, reveal global action periodicity and remain invariant to temporal locations.
 
\begin{figure}[t]
    \centering
    \includegraphics[width=0.5\linewidth]{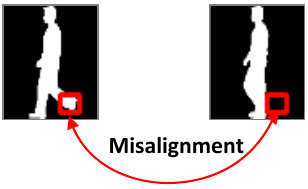}
    \caption{Illustration of the misalignment in gait sequences. Since pixels of the same spatial locations in different frames may correspond to different semantic contents, the utilization of temporal operations could lead to blurred or overlapped appearances.}
    \label{misalignment}
\end{figure}
 
 Next, a local relation modeling among these temporal features guides the network to adaptively enhance or suppress temporal features with different scales in each frame. Afterwards, a global relation modeling is involved to interact the multi-scale features along the whole sequence, constructing global communication to capture the most discriminative representation. Inspried by some recent works \cite{zhang2021multi, chen2021crossvit, dosovitskiy2020image, liu2023transkeleton, zhu2022mlst} that use Transformer \cite{transformer} to improve the model's ability to learn global relations, we adopt a Transformer block to model global relations across multi-scale features. Our method forms a hierarchical framework to adaptively aggregate multi-scale features in both local and global aspects, allowing for the modeling of complex motion and making it highly suitable for gait recognition.

 Furthermore, we notice a misalignment problem in the temporal modeling of gait recognition that has not yet been explored. As shown in Figure. \ref{misalignment}, the same pixel locations from different frames may correspond to different foregrounds and backgrounds. The utilization of temporal operations, such as temporal convolution and temporal pooling, naturally leads to blurry and corrupted appearances. However, appearance features could provide some auxiliary information for distinguishing different people. To address such issue, we propose a salient spatial feature learning (\textbf{SSFL}) module to select discriminative local spatial parts across the whole sequence, which can be considered as supplements to remedy the corruption in appearance features. The discriminative feature selection relies on evaluating the importance of temporal features, for which we employ the multi-head self-attention (MHSA) mechanism to generate groups of global importance maps. In particular, as described in \cite{transformer}, each head concentrates on a specific representation subspace, enabling multiple heads to generate diverse importance maps. These maps enable the selection of groups of discriminative local components from various perspectives.
 
 The adaptive temporal modeling and salient spatial learning provide complementary properties for each other. On one hand, adaptive temporal aggregation (ATA) module mainly considers temporal modeling and salient spatial feature learning (SSFL) module focuses on spatial learning. Specifically, ATA produces temporal aggregation of multi-scale clues which describes motion patterns, and SSFL selects well-perserved spatial features based on saliency evaluations, which are related to static images. On the other hand, MCAT aggregates temporal clues in a soft-attention way and SSFL obtains salient spatial features in a hard-attention manner. In a nutshell, by jointly investigating motion learning and spatial mining simultaneously, we achieve outstanding performance over the state-of-the-art methods.
 
The main contributions of the proposed method are summarized as the following three aspects:
\begin{itemize}
\item[$\bullet$] In this paper, we propose a temporal modeling network MCAT to fuse multi-scale temporal features in an adaptive way from both local and global aspects, which considers the cross-scale contextual information as a guidance for temporal aggregation. 
   
\item[$\bullet$] We propose a salient spatial feature learning (SSFL) module to remedy the feature corruption caused by temporal operations. SSFL constructs global feature importance maps to select salient spatial features across the whole sequence, which provide high quality spatial clues.

\item[$\bullet$] Extensive experiments conducted on three datasets, i.e., CASIA-B \cite{yu2006framework}, OU-MVLP \cite{takemura2018multi} and GREW \cite{grew}, demonstrate the state-of-the-art performance of MCAT. And further ablation experiments prove the effectiveness of the proposed modules. Additional experiments using practical settings reveal the real-world application potentials of MCAT. 
\end{itemize}
 
A preliminary conference version of this work was published in ICCV-2021 \cite{cstl}. We make three improvements to this work: (1) We extend the adaptive multi-scale feature aggregation from a local approach to a hierarchical approach that encompasses both local and global scales. This modification results in improved recognition performance.  (2) The former SSFL constructs importance maps on each frame individually, and only generates a group of salient parts. Instead, the improved SSFL forms the importance maps in a global view by utilizing the multi-head self-attention mechanism, which is capable of generating more groups of salient spatial parts, thereby enhancing spatial mining capacity. (3) We conduct additional experiments on a recently published large-scale gait dataset captured in the wild, i.e., GREW \cite{grew}, on which we further validate the effectiveness and robustness of our method. And more ablation experiments are conducted on CASIA-B to analyze the contribution of the proposed modules, explore the network design, and study the performances of our method in simulated real-world scenarios.

The rest of this paper is organized as follows. Firstly, the Section \ref{related_work} describes the related work. Next, the Section. \ref{proposed_method} gives the details of the proposed method. Afterwards, the Section. \ref{experiments_and_discussion} provides comprehensive experiments and corresponding analysis. Finally, the Section. \ref{conclusion} concludes the paper and present future work.

\section{Related Work}
\label{related_work}
\subsection{Gait Recognition}
Current gait recognition methods could be categorized into two types: model-based and appearance-based.

\textbf{Model-based} methods \cite{liao2020model, liao2017pose, teepe2021gaitgraph, huang2023condition} were proposed to model walking patterns and body structures of humans based on extracted key points \cite{cao2019openpose, sun2019deep, cao2017realtime}. Model-based methods are robust to variations of clothing and camera viewpoints. However, due to (1) inaccurate pose results estimated from low-quality images and (2) the missing of identity-related shape information, model-based methods are usually inferior to appearance-based methods in performance comparison.
 
\textbf{Appearance-based} methods \cite{chao2021gaitset, fan2020gaitpart, hou2020gait, lin2020gait, zhang2019cross, wolf2016multi, han2005individual, he2018multi, wu2016comprehensive, hu2013view} extracted spatio-temporal features binary silhouettes by CNN networks or handcrafted algorithms. Gait Energy Image (GEI) \cite{han2005individual} was generated through projecting a sequence of gait silhouettes into a single image. The GEI-based methods \cite{han2005individual, he2018multi, wu2016comprehensive, hu2013view, li2020gait, xu2020cross} greatly compressed computational cost but lost discriminative expression. In contrast, the video-based approaches \cite{chao2021gaitset, fan2020gaitpart, hou2020gait, lin2020gait, zhang2019cross, zhang2020learning, GaitGL, local3D, cui2022gaittransformer, huang2022star} processed gait sequences frame by frame, which maintained the frame-level discriminative feature in a large extent, and benefited the networks to learn temporal representation. Our approach belongs to appearance-based method and takes silhouette sequences as input.
\subsection{Temporal Modeling}
Current literatures proposed different strategies for gait temporal modeling, including 1D convolutions, LSTMs and 3D convolutions etc. Particularly, GaitSet \cite{chao2021gaitset}, GLN \cite{hou2020gait} and GaitBase \cite{fan2023opengait} considered a gait sequence as an unordered set, which mainly focused on spatial modeling and captured inter-frame dependency implicitly. GaitPart \cite{fan2020gaitpart} and Wu et al. \cite{wu2020condition} extracted local temporal clues by convolutions and aggregated them in a summation or a concatenation manner. LSTM networks were applied in \cite{zhang2019cross, zhang2020learning} to achieve long-short temporal modeling, which fused temporal clues by temporal accumulation. With the help of stacked 3D blocks, MT3D \cite{lin2020gait} and GaitGL \cite{GaitGL} incorporated temporal information with small and large scales, then concatenated or summed these features as outputs. 3DLocal \cite{local3DCNN} applied 3D CNN to obtain different local parts, and fused them with feature concatenation. GaitTransformer \cite{cui2022gaittransformer} proposed Multiple-Temporal-Scale Transformer (MTST) module to model the long-term features. However, current methods have obvious shortcomings in learning flexible and robust multi-scale temporal features, which are incapable of satisfying temporal modeling requirements for different body parts of gait motion. 
 
Recently, transformers were broadly applied in various computer vision tasks \cite{vit, swin, zhao2022spatial, tang2022multi}. Compared with the prevalent CNNs, the most outstanding strength of transformers is the global reasoning capacity, which empowers models to capture features in the long range. In the domain of video-based tasks, TPT \cite{zhang2023end}, ViViT \cite{vivit} and Vidtr \cite{vidtr} proved transformers could extract global temporal features effectively. 

Our method differs from above methods in three aspects: (1) MCAT generates temporal features in three scales, i.e., frame-level, short-term and long-term. Such rich temporal clues enable our network to obtain diverse motion learning ability. (2) For fusing multi-scale temporal features adaptively, MCAT employs both local and global cross-scale relation modelings to obtain the most appropriate temporal expression. (3) The transformer block we use is not for global feature extraction like current methods, but for global feature interaction on multiple time scales.
\subsection{Spatial Preserving}
The spatial appearance features provide supplementary cues to recognize people besides motion pattern. GaitNet \cite{zhang2020learning} proposed a disentanglement-based scheme to obtain the canonical feature to help recognition, whose goal was close to ours. Except for gait recognition, the spatial misalignment also degraded performance in other person-related recognition task, e.g. Person Re-identification. In video-based Person Re-identification, different methods were proposed to maintain the clearness of spatial features. In AP3D \cite{gu2020appearance}, researchers proposed Appearance Preserving Module (APM) to mitigate the misalignment problem in temporal modeling. APM used a feature similarity calculation strategy to match the foregrounds in continuous frames within a local window based on the color, texture and illumination et al. Chen et al. \cite{chen2020temporal} proposed a method dubbed Adversarial Feature Augmentation (AFA) to capture motion coherence by a adversarial form. However, AFA only employed motion-irrelevant features, but totally abandoned temporal clues.
 
Unfortunately, the above spatial preserving methods are not appropriate for gait recognition. GaitNet and APM are both operated on RGB-based appearance features, e.g., color, texture and illumination. But the inputs of our model are sequences of binary silhouettes, which do not include the needed appearance features. Moreover, AFA only incorporated the modeling of motion-irrelevant features but ignored the motion-relevant features, which are vital for gait recognition. 

Different from these strategies, in our approach, SSFL selects discriminative local parts to maintain the spatial characteristics of subjects, which is feasible for binary inputs. And this operation is parallel to temporal modeling, thus would not compromise temporal feature extraction.

\section{Proposed Method}
\label{proposed_method}
In this section, we firstly describe the overall pipeline of our method, then illustrate the detailed structure of key components in the network.

\subsection{Overall Architecture}
\label{overall_architecture}
\begin{figure*}[t]
\centering
\includegraphics[width=\linewidth]{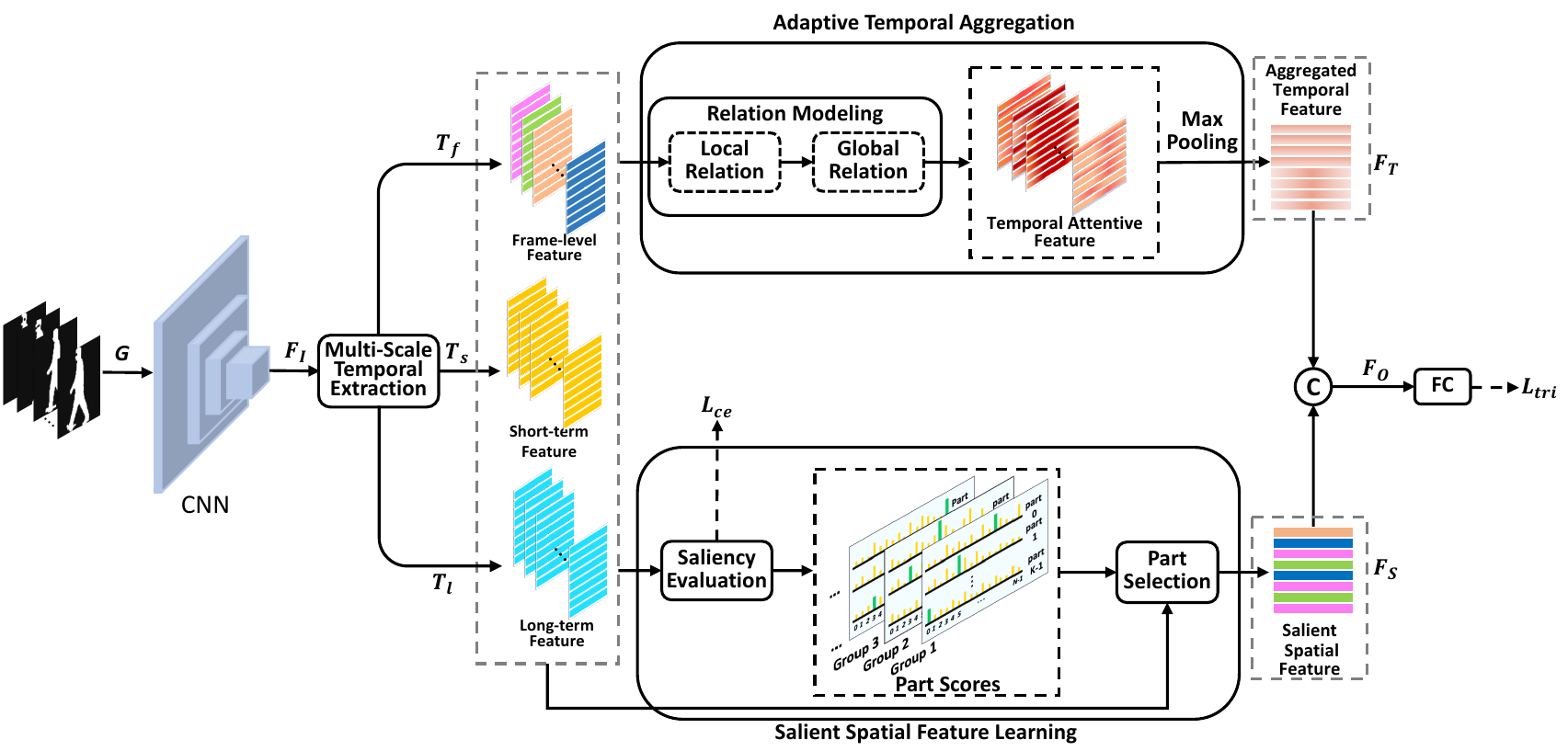}
\caption{Overview of MCAT. A sequence of gait silhouettes are firstly fed into a 2D CNN to extract spatial features in each frame. Then, a multi-scale temporal extraction (MSTE) module is utilized to obtain temporal features in three scales. Afterwards, a two-branch architecture composed of an adaptive temporal aggregation (ATA) module and a salient spatial feature learning (SSFL) module is formed to aggregate multi-scale features and select salient spatial parts respectively.  Arrows, $G$, $F_I$, $F_T$, $F_S$ and $F_O$ denote operations, input gait sequence, features processed by CNN, temporal aggregated features, salient spatial features and output features respectively. $L_{tri}$ and $L_{ce}$ represent triplet loss and cross-entropy loss respectively.}
\label{frame_work}
\end{figure*}

The overview structure of our method is presented in Figure. \ref{frame_work}. A batch of $B$ gait samples of $N$ frames are fed into the network as input, which is denoted as $G \in \mathbb{R} ^ {B \times N \times H \times W}$ where $H$ and $W$ denote the height and width of each frame respectively. Firstly, $G$ is passed through a 2D CNN to extract spatial feature $F_I \in \mathbb{R} ^ {B \times N \times C \times H/2 \times W/2}$, where $C$ denotes the number of feature channels. The detailed CNN architecture is given in Table. \ref{tab:backbone}. Afterwards, we implement a multi-scale temporal extraction (MSTE) module on $F_I$ to generate temporal features with three different temporal scales, i.e., frame-level, short-term and long-term, which are denoted as $T_f$, $T_s$ and $T_l$ respectively. $T_f$, $T_s$ and $T_l$ are all with size of $\mathbb{R} ^ {B \times N \times C \times K}$, where $K$ denotes the number of horizontal divided feature parts that correspond to body parts in some extent. Next, temporal features are fed into adaptive temporal aggregation (ATA) and salient spatial feature learning (SSFL) blocks respectively, through which we obtain temporal aggregated feature $F_T \in \mathbb{R} ^ {B \times C \times K}$ and salient spatial feature $F_S \in \mathbb{R} ^ {B \times C \times K}$ correspondingly. Temporal aggregated feature $F_T$ is a temporal summarization of the whole sequence, which represents the discriminative information in temporal domain. Spatial salient feature $F_S$ is obtained by selecting groups of salient spatial parts, which maintain rich high-quality silhouette information. Finally, $F_S$ and $F_T$ are concatenated along channel dimension as outputs $F_O$.

\begin{table}[ht]
 \centering
 \caption{Structure of the backbone on CASIA-B. $C_{in}$ and $C_{out}$ denote the input channel and output channel of each layer respectively.}
 \begin{tabular}{|c|c|c|c|c|c|}
 \hline
    Layer & $C_{in}$ & $C_{out}$ & Kernel & Pad & Activation  \\\hline
    Conv1 & 1 & 32 & 3 & 1 & Leaky ReLU \\\hline
    \multirow{2}{*}{Conv2} & 32 & 64 & 3 & 1 & Leaky ReLU \\\cline{2-6}
    & \multicolumn{5}{c|}{Max Pooling kernel=(2,2), stride=2} \\\hline
    Conv3 & 64 & 128 & 3 & 1 & Leaky ReLU\\\hline
    Conv4 & 128 & 128 & 3 & 1 & Leaky ReLU\\\hline
 \end{tabular}
 \label{tab:backbone}
\end{table}

\begin{figure}[t]
     \centering
     \includegraphics[width=0.6\linewidth]{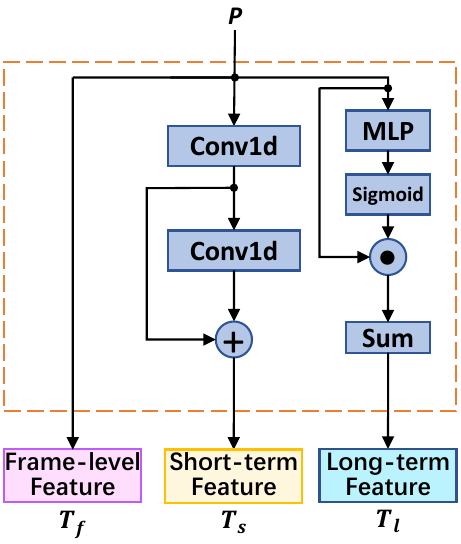}
     \caption{Detailed architecture of Multi-Scale Temporal Extraction, which produces temporal features in three levels for each frame.}
     \label{MTFL}
 \end{figure}

\subsection{Multi-Scale Temporal Extraction}
As discussed in Section. \ref{overall_architecture}, we aim to enrich the diversity of temporal features. Firstly, we divide $F_I$ into $K$ parts vertically, then apply Global Max Pooling (GMP) and Global Average Pooling (GAP) to obtain part-level pooling features $P \in \mathbb{R} ^ {B \times N \times C \times K}$, where $P^{b,n}$ represents part-level features of the $n$-th frame in the $b$-th sample. As shown in Figure. \ref{MTFL}, the frame-level features are the duplicate of $P$, which do not get involved with temporal operation, thus the appearance characteristics of each time instant are well-maintained.

In order to capture short-term temporal features, we apply two serial 1D convolutions with kernel size of 3, and add the features after each 1D convolution as $T_s$. Obtaining short-term features enables the network to focus on short period temporal motion patterns and subtle changes with perceptive fields of 3 and 5.

The long-term feature extraction is based on the combination of all frames. Firstly, a Multi-layer Perceptron (MLP) followed by a Sigmoid function is applied on $P$ to evaluate the importance of different frames. Next, the weighted summation of all frames by the importance scores is utilized as the long-term temporal features $T_l$, which is formulated as:
\begin{equation}
 T_l^b = \frac{\sum_{n=1}^N{Sigmoid(MLP(P^{b,n}))} \odot P^{b,n}}{\sum_{n=1}^N{Sigmoid(MLP(P^{b,n}))}},
\end{equation}

\noindent where $\odot$ denotes dot product. It should be noted that, $T_l^b$ is invariant for all frames in the $b$-th sample, which describes global motion cues. After that, we obtain temporal features of three levels, e.g., $T_f$, $T_s$ and $T_l$, for subsequent ATA and SSFL blocks.
 
 \subsection{Adaptive Temporal Aggregation}
\begin{figure}[t]
    \centering  %居中
    \subfigure[Local relation modeling module.]{ %第一张子图
    \centering
    \begin{minipage}{8cm}
    \centering    %子图居中
    \includegraphics[scale=0.55]{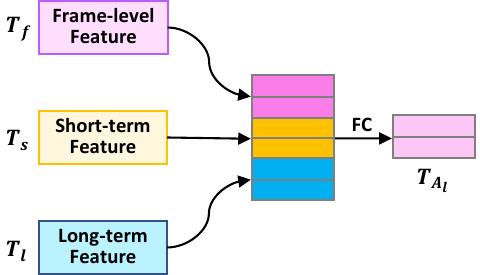}%以pic.jpg的0.5倍大小输出
    \end{minipage}
    \label{fig:local_relation}
    }
    \subfigure[Global relation modeling module.]{ %第二张子图
    \centering
    \begin{minipage}{8cm}
    \centering    %子图居中
    \includegraphics[width=\linewidth]{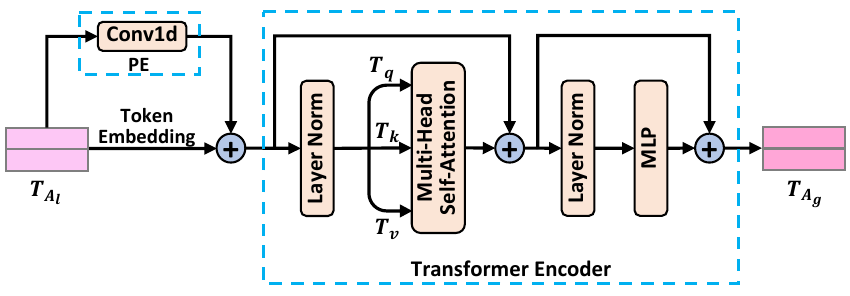}
    \end{minipage}
    \label{fig:global_relation}
    }
    \caption{The detailed structure for modeling local-global relations between multi-scale features.}
    \label{fig:local_global_relation}    %图片引用标记
\end{figure}

 In this part, we utilize multi-scale temporal features to explore feature relations, which enable information exchanging among different temporal scales. As discussed in \cite{fan2020gaitpart}, different body parts own various motion patterns, which indicates the diverse expressions are needed for temporal modeling. Intuitively, as shown in Figure. \ref{fig:local_global_relation}, the interaction of different scales of features would effectively enrich the diversity of temporal representation, thus produce suitable motion expression for human body. As stated in Section. \ref{sec:introduction}, the feature interactions are conducted both locally and globally, which are illustrated below.
 
 \subsubsection{Local Relation Modeling}
 % We devise three strategies to achieve local aggregation in each frame. 
 
 % (1) We use a max pooling to obtain the most discriminative clues of the three scales, which is formulated as:
 % \begin{equation}
 %     T_{A_l}^{b,n} = Max(T_{f}^{b,n}, T_{s}^{b,n}, T_{l}^{b,n}),
 % \end{equation}
 % where $T_{A_l}^{b,n} \in \mathbb{R}^{C \times K}$ represents the local aggregated features in the $n$-th frame of the $b$-th sample.
 
 In order to achieve local aggregation in each frame, we concatenate the features from three scales along the channel dimension. Then we take a fully-connected (FC) layer to correlate the channels and obtain the fused output, which is represented as:
 \begin{equation}
     T_{A_l}^{b,n} = FC(T_{f}^{b,n} \textcircled{c} T_{s}^{b,n} \textcircled{c} T_{l}^{b,n}),
 \end{equation}
 where $\textcircled{c}$ denotes the concatenation operation and FC is a 1D depth-wise convolutions with kernel size of 1.

 Subsequently, the local aggregated feature $T_{A_l}$ is utilized as input to the global relation modeling module.

\subsubsection{Global Relation Modeling}
 To achieve global interactions, we use transformer to construct multi-scale relation modeling across the whole sequence. As discussed in \cite{transformer, vit, swin}, position encoding (PE) plays an important role to achieve permutation-variant modeling in transformers. Considering the various lengths of gait sequences, we adopt conditional position encoding\cite{CondTrans} to extract PE, which can fit the sequence length adaptively. As shown in Figure. \ref{fig:global_relation}, PE is obtained by applying 1D depth-wise convolutions with kernel size of 3 on $T_{A_l}$, which produces feature $T_{trans} \in \mathbb{R}^{B \times K \times N \times C}$. Following \cite{transformer}, we first apply a layer normalization on $T_{trans}$ and then conduct three feature transformations with three learnable matrix weights $W_q$, $W_k$ and $W_v$, which are formulated as:
 %  \begin{figure}[t]
 %     \centering
 %     \includegraphics[width=0.7\linewidth]{fig/MHSA_SSFL.pdf}
 %     \caption{The detailed structure of global relation modeling of multi-scale features using the transformer architecture. Particularly, the depth-wise (DW) conv is used to learn the position encoding conditioned on the input features dynamically.}
 %     \label{fig:global relation modeling}
 % \end{figure}
 \begin{equation}
    \begin{aligned}
    T_{norm} = Norm(T_{trans})
    \end{aligned}
    \label{eq:layernorm}
 \end{equation}
 \begin{equation}
     \begin{aligned}
     T_q = T_{norm}W_q, \quad T_k = T_{norm}W_k, \quad T_v = T_{norm}W_v,
    %  &\widetilde{T}_f = T_f \\
    %  &\widetilde{T}_s = T_f + T_s \\
    %  &\widetilde{T}_l = T_f + T_s + T_l.
     \end{aligned}
     \label{eq:transformation}
 \end{equation}
 where $W_q$, $W_k$ and $W_v$ are all with the dimension of $N_{head} \times C \times C/N_{head}$, and $N_{head}$ denotes the number of heads. Then, the attention matrix is obtained by multiplying $T_q$ with $T_k$ followed by a Softmax normalization:
 \begin{equation}
     A_T = Softmax(T_qT_k/\sqrt{C}),
     \label{eq:transformer_aggeragation}
 \end{equation}
 where $A_T \in \mathbb{R}^{B \times N_{head} \times K \times N \times N}$ and $\sqrt{C}$ is used to normalize the dot-product scores. Afterwards, we utilize the attention matrix $A_T$ to generate attentive feature $T_a$ with a shortcut and a normalization layer:
 \begin{equation}
    T_a = A_TT_v + T_{trans},
 \end{equation}
 where $T_a \in \mathbb{R}^{B \times K \times N \times C}$. Subsequently, we use a layer normalization and a feed-forward network (FFN) \cite{transformer}, such as MLP, to obtain the globally fused multi-scale feature $T_{A_g}$:
 \begin{equation}
     T_{A_g} = FFN(Norm(T_a))+T_a,
 \end{equation}
 where $T_{A_g} \in \mathbb{R}^{B \times K \times N \times C}$. Finally, a max pooling along temporal dimension is applied on $T_{A_g}$ to obtain sequence-level temporal representation $F_T \in \mathbb{R}^{B \times K \times C}$.
 
% \begin{figure}[t]
%      \centering
%      \includegraphics[width=\linewidth]{fig/local_global_relation_modeling.pdf}
%      \caption{The detailed structure for modeling local-global relations between multi-scale features.}
%      \label{fig:local_global_relation}
%  \end{figure}

 \subsection{Salient Spatial Feature Learning}
In this section, we aim to extract salient spatial parts to mitigate the damage in appearance features.
 \subsubsection{Discussion} 
\begin{figure}[t]
    \centering  %居中
    \subfigure[Select a discriminative frame across the sequence.]{   %第一张子图
    \begin{minipage}{8cm}
    \centering  %子图居中
    \label{fig:select frame}
    \includegraphics[scale=0.55]{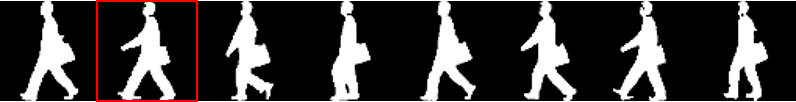}  %以pic.jpg的0.5倍大小输出
    \end{minipage}
    }
    
    \subfigure[Select the salient parts individually across the sequnece.]{ %第二张子图
    \begin{minipage}{8cm}
    \centering    %子图居中
    \label{fig:select parts}
    \includegraphics[scale=0.55]{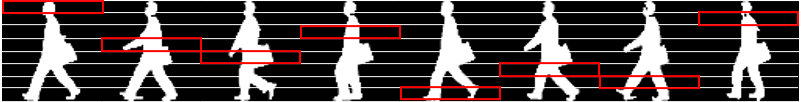}%以pic.jpg的0.5倍大小输出
    \end{minipage}
    }
    \caption{Illustration of two ways to select salient spatial features. Compared with selecting a frame, selecting the salient parts is conducted in a more fine-grained level, thus could obtain more high-quality local features.}    %大图名称
    \label{fig:illustration_SSFL}    %图片引用标记
\end{figure}

Intuitively, in order to remedy the corrupted spatial features, we should select an individual frame as the methods in \cite{gowda2020smart, kopuklu2019you}, which is illustrated in Figure. \ref{fig:select frame}. However, due to the various camera viewpoints, motion occlusions and imperfect segmentations, a single frame is probably incapable of expressing appearance features for all body parts clearly. Actually, the high quality body parts appear and disappear from frame to frame. Therefore, by utilizing such inherent motion characteristics, we select salient body parts across the whole sequence. As shown in Figure. \ref{fig:select parts}, we obtain local discriminative features in different frames instead of directly selecting one frame.
 
 Since the transformer is able to correlate global information based on the attention matrix adaptively, the weights in the attention matrix could reveal feature importance in some extent. Naturally, a body part that contains clearer feature representation has a stronger weight than a body part with vague feature representation. Therefore, we consider using the attention matrix as a metric to select salient spatial parts. Besides, due to the diversities in multiple heads\cite{transformer}, attention matrices in different heads may focus on informative features from various aspects, which enhances the spatial mining capacity.
 
 \subsubsection{Operation} Considering temporal clues provide contextual information for evaluating the discrimination of each frame, we firstly utilize a FC to fuse the multi-scale features. Then, we construct the attention matrix $A_s \in \mathbb{R}^{B \times N_{head} \times K \times N \times N}$ as in Equation. \ref{eq:transformation} and Equation. \ref{eq:transformer_aggeragation}, but remove the scale operation and Softmax function. Next, we squeeze the attention matrix along the column dimension, which is formulated as:
 \begin{equation}
     \widetilde{A}_s^{b,:,:,:} = \sum_{n=1}^{N}{A_s^{b,:,:,:,n}},
     \label{eq:matrix_squeeze}
 \end{equation}
 where $\widetilde{A}_s^{b,:,:,:} \in \mathbb{R}^{N_{head} \times K \times N}$ denotes the part scores of $K$ parts from $N_{head}$ heads in the $b$-th sample. The values of part scores represent feature importance, thus higher scores indicate clearer spatial representation. In order to supervise the correctness of part scores, we enforce a cross-entropy loss on the weighted summation of $T_f$ and $\widetilde{A}_s$. Here, the weighted feature $F_w$ is obtained by:
%  we firstly use a FC layer to transform the weighted part features, which can be represented as:
 
  \begin{equation}
     F_w^b = \sum_{n=1}^N{T_f^{b,n} \odot \widetilde{A}_s^{b,:,:,n}},
 \end{equation}
 where $F_w^b \in \mathbb{R}^{N_{head} \times K \times C}$ of the $b$-th sample.
 Then, a FC layer is utilized to transform $F_w$ into classification logics $P_w \in \mathbb{R}^{B \times N_{head} \times K \times C_t}$, where $C_t$ denotes the number of training subjects. Afterwards, the cross-entropy loss is applied on $P_w$ to produce $L_{ce}$:
 \begin{equation}
     L_{ce} = -\sum_{b=1}^B \sum_{h=1}^{N_{head}} \sum_{c=1}^{C_t} y_{b,c}\log(SoftMax(P_w^{b,h}))_c  ,
 \end{equation}
 where $y_{b,c}$ indicates the identity information of the $b$-th sample, which equals 0 or 1.
 Next, we obtain part indexes of the highest scores along temporal dimension:
 \begin{equation}
    x^{b,h,k} = \arg\max_{n}{\widetilde{A}_s^{b,h,k,n}},
 \end{equation}
 where $x^{h,b,k}$ denotes the temporal index of the selected $k$-th part in the $h$-th head of the $b$-th sample. By using the index $\{x^{b,h,k}|k=1,2,...,K\}$ in a hard-attention manner, we obtain the recombinant frame feature $F_r$ in the $h$-th head of the $b$-th sample:
 \begin{equation}
     \begin{aligned}
     F_r^{b,h} = T_f^{b,x^{b,h,1}} \textcircled{c} T_f^{b,x^{b,h,2}} \cdots \textcircled{c} T_f^{b,x^{b,h,k}},
     \end{aligned}
 \end{equation}
 where $\textcircled{c}$ denotes concatenation along part dimension. In the end, we fuse the recombinant features $F_r$ with the weighted features $F_w$ in different heads by:
 \begin{equation}
     \begin{aligned}
     F_S^{b} = (\sum_{h=1}^{N_{head}}{F_r^{b,h}}) \textcircled{c} (\sum_{h=1}^{N_{head}}{F_w^{b,h}}),
     \end{aligned}
 \end{equation}
 where $F_S^b \in \mathbb{R}^{K \times C}$ denotes the obtained salient spatial features of the $b$-th sample, and $\textcircled{c}$ denotes channel concatenation. $F_S$ offers supplementary spatial clues for temporal aggregated features $F_T$. Triplet loss \cite{hermans2017defense} is employed on the combination of $F_S$ and $F_T$ as metric learning loss function. The overall loss function is presented as following:
 
 \begin{equation}
    L = L_{ce} + L_{tri}
 \end{equation}
 
\section{Experiments and Discussion}
\label{experiments_and_discussion}
 \subsection{Datasets and Evaluation Metrics}
 We conduct experiments on three standard datasets, i.e., CASIA-B \cite{yu2006framework}, OU-MVLP \cite{takemura2018multi} and GREW \cite{grew}, to verify the superiority of our method. Further ablation experiments are conducted on CASIA-B to demonstrate the positive impact of each component in our method.
 
\noindent \textbf{CASIA-B.} CASIA-B \cite{yu2006framework} is composed of 124 subjects, and each subject contains 110 sequences with 11 different camera views. Under each camera view, every subject contains three walking conditions, i.e., normal (NM) (6 sequences), walking with bag (BG) (2 sequences) and walking with coat (CL) (2 sequences). For the training and testing stages, we follow the protocols in \cite{wu2016comprehensive}. The samples from the first 74 subjects are considered as training set, and the remaining 50 subjects are considered as test set. At testing phase, the first 4 sequences in NM condition of each subject are regarded as gallery set and the remaining 6 sequences of each subject are used as probe set, including 2 sequences of NM, 2 sequences of BG and 2 sequences of CL.

\noindent \textbf{OU-MVLP.} OU-MVLP \cite{takemura2018multi} is composed of 10307 subjects. Each subject contains 28 sequences with 14 camera views, thus each subject contains 2 sequences (index '01' and '02') for each view. The first 5153 subjects are used for training, while the remaining 5154 subjects are for testing. In particular, the sequences with index '01' are regarded as gallery and the sequences with index '02' are regarded as probe set at testing phase.

\noindent \textbf{GREW.} GREW \cite{takemura2018multi} captures gait videos under uncontrolled conditions, which is composed of 26345 subjects and, 128671 sequences in total. In particular, 882 cameras are involved to record the videos, where the corresponding views are not predefined like CASIA-B or OU-MVLP. Concretely, the training set includes 102887 sequences of 20000 subjects, the validation set includes 1784 sequences of 345 subjects and the testing set includes 24000 sequences of 6000 subjects. 

\subsection{Implementation Details}
% \noindent \textbf{Measurement of Human Focus Distribution.} The human focus distribution in Figure.\ref{fig:motivation} is a qualitative illustration of how people distinguish different sequences. Seven volunteers were asked to observe ten misclassified sequences carefully to determine which frames could be used to distinguish them. The focus distribution is obtained according to the voting results. Since this process is quite time-consuming, we did not conduct it on whole datasets. But we believe the conclusion is correct and consistent with life experience and intuition.
\subsubsection{Hyper-parameters}
(1) Follow the settings in \cite{cstl,GaitGL}, we set the value of $B$ (number of training samples in one iteration) as 64, 256 and 256 on CASIA-B \cite{yu2006framework}, OU-MVLP \cite{takemura2018multi} and GREW \cite{grew} datasets respectively. (2) The value of $N$ (input frame number) and $K$ (part division number) are set as 30 and 32. (3) The number of output channels for FC shown in Figure. \ref{frame_work} is set to 256, 512 and 512 for CASIA-B \cite{yu2006framework}, OU-MVLP \cite{takemura2018multi} and GREW \cite{grew} datasets respectively. (4) All MLPs follow: FC(c, c/16)-$>$ReLU()-$>$FC(c/16, c). The two FCs in ATA are FC(c, c/16) and FC(c/16, c).
\subsubsection{Training Details}
(1) Each frame is aligned as \cite{takemura2018multi} does, and we resize each frame to the size of $64 \times 44$ or $128 \times 88$. For each input sequence, we follow the frame sampling strategy as \cite{fan2020gaitpart} does. (2) We apply separate Batch All ($BA_+$) triplet loss to train our network. The batch size for training is noted as $(p, k)$, where $p$ denotes the number of sampled subjects and $k$ denotes the number of sampled sequences for each subject. Particularly, $(p, k)$ is set to $(8, 8)$ on CASIA-B and $(32, 8)$ on OU-MVLP and GREW. (3) The backbone architecture on CASIA-B is given in Table. \ref{tab:backbone}. Since the data amount of OU-MVLP and GREW is much larger than that of CASIA-B, the numbers of output channels of each layer in Table. \ref{tab:backbone} are set to 64, 128, 256, 256 on OU-MVLP and GREW datasets, which follows the design in GaitSet \cite{chao2021gaitset}. (4) Totally, we train 120k iterations on CASIA-B and 250k iterations on OU-MVLP and GREW. Morever, our model is optimized by Adam, and the learning rate is started to set as 1e-4 and reduced to 1e-5 at 150k iterations on OU-MVLP and GREW. We implement our models on Pytorch \cite{paszke2017automatic} platform and use four NVIDIA GeForce GTX 3090 GPUs to perform our experiments.

\begin{table*}[t]
\begin{center}
\footnotesize
\caption{Averaged rank-1 accuracies (\%) on CASIA-B, excluding identical-view cases. Std denotes the performance sample standard deviation across 11 views.}
\begin{tabular}{|c|c|c| l l l l l l l l l l l|l|c|}
\hline
\multicolumn{2}{|c|}{Gallery NM} & Resolution & \multicolumn{11}{c|}{$0-180^\circ$} & \multirow{2}{*}{Mean} & \multirow{2}{*}{Std} \\\cline{1-14}
\multicolumn{2}{|c|}{Probe} & $-$ & $0^\circ$ & $18^\circ$ & $36^\circ$ & $54^\circ$ & $72^\circ$ & $90^\circ$ & $108^\circ$ & $126^\circ$ & $144^\circ$ & $162^\circ$ & $180^\circ$ & & \\\hline
\multirow{7}{*}{NM} & GaitSet\cite{chao2021gaitset} & $64 \times 44$ & 91.1 & 99.0 & 99.9 & 97.8 & 95.1 & 94.5 & 96.1 & 98.3 & 99.2 & 98.1 & 88.0 & 96.1 & 3.5\\
& GaitPart \cite{fan2020gaitpart} & $64 \times 44$ & 94.1 & 98.6 & 99.3 & 98.5 & 94.0 & 92.3 & 95.9 & 98.4 & 99.2 & 97.8 & 90.4 & 96.2 & 3.1 \\
% & MT3D \cite{lin2020gait} &$64 \times 44$ & 95.7 & 98.2 & 99.0 & 97.5 & 95.1 & 93.9 & 96.1 & 98.6 & 99.2 & 98.2 & 92.0 & 96.7 & 2.3 \\
& GaitGL \cite{GaitGL} & $64 \times 44$ & 96.0 & 98.3 & 99.0 & 97.9 & 96.9 & 95.4 & 97.0 & 98.9 & 99.3 & 98.8 & 94.0 & 97.4 & 1.7 \\
& 3DLocal \cite{local3D} & $64 \times 44$ & 96.0 & 99.0 & \textbf{99.5} & 98.9 & 97.1 & 94.2 & 96.3 & 99.0 & 98.8 & 98.5 & 95.2 & 97.5 & 1.8 \\
% & RPNet \cite{qin2021rpnet} & $64 \times 44$ & 95.1 & 99.0 & 99.1 & 98.3 & 95.7 & 93.6 & 95.9 & 98.3 & 98.6 & 97.7 & 90.8 & 96.5 & 2.5 \\
& GaitTransformer \cite{cui2022gaittransformer} & $64 \times 44$ & 94.9 & 98.3 & 98.4 & 97.8 & 94.8 & 94.1 & 96.3 & 98.5 & 99.0 & 98.3 & 90.7 & 96.5 & 2.5 \\
% & Huang et al. \cite{huang2022enhanced} & $64 \times 44$ & 95.6 & 98.6 & 99.1 & 97.9 & 96.7 & 94.4 & 96.9 & 98.7 & 99.3 & 98.6 & 95.1 & 97.4 & 1.6 \\
& Lagrange \cite{chai2022lagrange} & $64 \times 44$ & 95.2 & 97.8 & 99.0 & 98.0 & 96.9 & 94.6 & 96.9 & 98.8 & 98.9 & 98.0 & 91.5 & 96.9 & 2.2 \\
& STAR \cite{huang2022star} & $64 \times 44$ & 96.5 & 98.7 & 99.0 & 97.6 & 96.1 & 95.4 & 96.8 & 98.6 & 99.1 & 98.9 & 93.9 & 97.3 & 1.6 \\
& GaitBase \cite{fan2023opengait} & $64 \times 44$ & - & - & - & - & - & - & - & - & - & - & - & 97.6 & - \\
% & \textbf{MCAT} (Ours) & $64 \times 44$ & \textbf{97.7} & \textbf{99.3} & 99.4 & \textbf{99.1} &\textbf{97.3} & \textbf{96.1} & \textbf{98.5} & \textbf{99.7} & \textbf{99.7} & \textbf{99.2} & \textbf{96.9} & \textbf{98.5} & \textbf{1.2} \\
& \textbf{MCAT} (Ours) & $64 \times 44$ & \textbf{97.7} & \textbf{99.3} & 99.4 & \textbf{99.1} &\textbf{97.3} & \textbf{96.1} & \textbf{98.5} & \textbf{99.7} & \textbf{99.7} & \textbf{99.2} & \textbf{96.9} & \textbf{98.5} & \textbf{1.2} \\\cline{2-16}\cline{2-16}

& GLN \cite{hou2020gait} & $128 \times 88$ & 93.2 & 99.3 & 99.5 & 98.7 & 96.1 & 95.6 & 97.2 & 98.1 & 99.3 & 98.6 & 90.1 & 96.9 & 3.0 \\
& 3DLocal \cite{local3D} & $128 \times 88$ & \textbf{97.8} & 99.4 & \textbf{99.7} & \textbf{99.3} & 97.5 & 96.0 & 98.3 & 99.1 & 99.9 & 99.2 & 94.6 & 98.3 & 1.7 \\
% & \textbf{MCAT} (Ours) & $128 \times 88$ & 97.6 & \textbf{99.4} & 99.3 & 98.9 & \textbf{97.9} & \textbf{97.9} & \textbf{98.9} & \textbf{99.8} & \textbf{99.9} & \textbf{99.6} & \textbf{96.7} & \textbf{98.7} & \textbf{1.0} \\\hline
& STAR \cite{huang2022star} & $128 \times 88$ & 95.8 & 98.9 & 99.0 & 98.0 & 96.0 & 94.4 & 96.8 & 98.9 & 99.3 & 99.4 & 94.4 & 97.4 & 1.8 \\
& \textbf{MCAT} (Ours) & $128 \times 88$ & 97.6 & \textbf{99.4} & 99.3 & 98.9 & \textbf{97.9} & \textbf{97.9} & \textbf{98.9} & \textbf{99.8} & \textbf{99.9} & \textbf{99.6} & \textbf{96.7} & \textbf{98.7} & \textbf{1.0} \\\hline

\multirow{7}{*}{BG} & GaitSet \cite{chao2021gaitset} & $64 \times 44$ & 86.7 & 94.2 & 95.7 & 93.4 & 88.9 & 85.5 & 89.0 & 91.7 & 94.5 & 95.9 & 83.3 & 90.8 & 4.4 \\
& GaitPart \cite{fan2020gaitpart} & $64 \times 44$ & 89.1 & 94.8 &96.7& 95.1& 88.3& 84.9 &89.0 &93.5 &96.1 &93.8& 85.8& 91.5 & 4.2 \\
% & MT3D \cite{lin2020gait} & $64 \times 44$ & 91.0 & 95.4 & 97.5 & 94.2 & 92.3 & 86.9 & 91.2 & 95.6 & 97.3 & 96.4 & 86.6 & 93.0 & 3.9 \\
& GaitGL \cite{GaitGL} & $64 \times 44$ & 92.6 & 96.6 & 96.8 & 95.5 & \textbf{93.5} & 89.3 & 92.2 & \textbf{96.5} & \textbf{98.2} & 96.9 & 91.5 & 94.5 & 2.8\\
& 3DLocal \cite{local3D} & $64 \times 44$ & 92.9 & 95.9 & \textbf{97.8} & \textbf{96.2} & 93.0 & 87.8 & \textbf{92.7} & 96.3 & 97.9 & \textbf{98.0} & 88.5 & 94.3 & 3.5 \\
% & RPNet \cite{qin2021rpnet} & $64 \times 44$ & 92.3 & 96.6 & 96.6 & 94.5 & 91.9 & 87.6 & 90.7 & 94.7 & 96.0 & 93.9 & 86.1 & 92.8 & 3.4 \\
& GaitTransformer \cite{cui2022gaittransformer} & $64 \times 44$ & 89.9 & 94.5 & 95.9 & 94.6 & 93.9 & 88.8 & 91.1 & 96.3 & 98.1 & 97.3 & 88.9 & 93.5 & 3.7 \\
% & Huang et al. \cite{huang2022enhanced} & $64 \times 44$ & 92.7 & 95.9 & 96.3 & 94.9 & 93.2 & 87.7 & 90.9 & 96.2 & 97.3 & 96.9 & 91.7 & 94.0 & 2.9 \\
& Lagrange \cite{chai2022lagrange} & $64 \times 44$ & 89.9 & 94.5 & 95.9 & 94.6 & 93.9 & 88.0 & 91.1 & 96.3 & 98.1 & 97.3 & 88.9 & 93.5 & 3.2 \\
& STAR \cite{huang2022star} & $64 \times 44$ & 92.3 & 96.7 & 97.1 & 95.6 & 92.6 & 88.5 & 92.3 & 96.0 & 97.0 & 95.7 & 89.0 & 93.9 & 3.0 \\
& GaitBase \cite{fan2023opengait} & $64 \times 44$ & - & - & - & - & - & - & - & - & - & - & - & 94.0 & - \\
% & \textbf{MCAT} (Ours) & $64 \times 44$ & \textbf{94.6} & \textbf{97.5} & 97.0 & 95.7 & 92.0 & \textbf{90.2} & 91.9 & 96.2 & 97.9 & 97.2 & \textbf{92.5} & \textbf{94.8} & \textbf{2.7} \\\cline{2-16}
& \textbf{MCAT} (Ours) & $64 \times 44$ & \textbf{94.6} & \textbf{97.5} & 97.0 & 95.7 & 92.0 & \textbf{90.2} & 91.9 & 96.2 & 97.9 & 97.2 & \textbf{92.5} & \textbf{94.8} & \textbf{2.7} \\\cline{2-16}

& GLN \cite{hou2020gait} & $128 \times 88$ & 91.1 & 97.7 & 97.8 & 95.2 & 92.5 & 91.2 & 92.4 & 96.0 & 97.5 & 95.0 & 88.1 & 94.0 & 3.2 \\
& 3DLocal \cite{local3D} & $128 \times 88$ & 94.7 & \textbf{98.7} & \textbf{98.8} & \textbf{97.5} & 93.3 & 91.7 & 92.8 & 96.5 & \textbf{98.1} & 97.3 & 90.7 & 95.5 & 2.9 \\
% & \textbf{MCAT} (Ours) & $128 \times 88$ &\textbf{95.9} & 97.1 & 97.8 & 97.2 & \textbf{95.1} & \textbf{93.0} & \textbf{96.1} & \textbf{97.5} & 98.0 & \textbf{97.4} & \textbf{93.8} & \textbf{96.2} & \textbf{1.7} \\\hline
& STAR \cite{huang2022star} & $128 \times 88$ & 93.6 & 98.0 & 98.3 & 96.1 & 93.9 & 91.7 & 94.8 & \textbf{97.6} & 98.2 & \textbf{97.5} & 91.8 & 95.6 & 2.4 \\
& \textbf{MCAT} (Ours) & $128 \times 88$ &\textbf{95.9} & 97.1 & 97.8 & 97.2 & \textbf{95.1} & \textbf{93.0} & \textbf{96.1} & 97.5 & 98.0 & 97.4 & \textbf{93.8} & \textbf{96.2} & \textbf{1.7} \\\hline

\multirow{7}{*}{CL} & GaitSet \cite{chao2021gaitset} & $64 \times 44$ & 59.5 & 75.0 & 78.3 & 74.6 & 71.4 & 71.3 & 70.8 & 74.1 & 74.6 & 69.4 & 54.1 & 70.3 & 7.2 \\
% &  & $128 \times 88$ & 66.3 & 79.4 & 84.5 & 80.7 & 74.6 & 73.2 & 74.1 & 80.3 & 79.7 & 72.3 & 62.9 & 75.3 \\\cline{2-15}
& GaitPart \cite{fan2020gaitpart} & $64 \times 44$ & 70.7 &85.5 &86.9 &83.3 &77.1 &72.5 &76.9 &82.2 &83.8 &80.2 &66.5 &78.7 &6.6 \\
% & MT3D \cite{lin2020gait} & $64 \times 44$ & 76.0 & 87.6 & 89.8 & 85.0 & 81.2 & 75.7 & 81.0 & 84.5 & 85.4 & 82.2 & 68.1 & 81.5 & 6.2 \\
& GaitGL \cite{GaitGL} & $64 \times 44$ & 76.6 & 90.0 & 90.3 & 87.1 & 84.5 & 79.0 & 84.1 & 87.0 & 87.3 & 84.4 & 69.5 & 83.6 & 6.3 \\
& 3DLocal \cite{local3D} & $64 \times 44$ & 78.2 & 90.2 & 92.0 & 87.1 & 83.0 & 76.8 & 83.1 & 86.6 & 86.8 & 84.1 & 70.9 & 83.7 & 6.2 \\
% & RPNet \cite{qin2021rpnet} & $64 \times 44$ & 75.6 & 87.1 & 88.3 & 83.1 & 78.8 & 78.0 & 79.9 & 82.7 & 83.9 & 78.9 & 66.6 & 80.3 & 5.7 \\
& GaitTransformer \cite{cui2022gaittransformer} & $64 \times 44$ & 81.5 & \textbf{91.9} & 92.2 & 91.2 & \textbf{85.9} & \textbf{83.1} & \textbf{86.8} & \textbf{90.7} & 90.4 & \textbf{89.0} & \textbf{75.6} & \textbf{87.1} & \textbf{5.0} \\
% & Huang et al. \cite{huang2022enhanced} & $64 \times 44$ & 75.6 & 89.2 & 92.4 & 90.3 & 84.3 & 80.2 & 83.0 & 86.3 & 89.0 & 83.9 & 69.8 & 84.0 & 6.4 \\
& Lagrange \cite{chai2022lagrange} & $64 \times 44$ & \textbf{81.6} & 91.0 & \textbf{94.8} & \textbf{92.2} & 85.5 & 82.1 & 86.0 & 89.8 & \textbf{90.6} & 86.0 & 73.5 & 86.6 & 5.7 \\
& STAR \cite{huang2022star} & $64 \times 44$ & 77.9 & 89.5 & 92.1 & 88.2 & 83.1 & 80.8 & 83.3 & 86.5 & 88.7 & 85.8 & 68.3 & 84.0 & 6.3 \\
& GaitBase \cite{fan2023opengait} & $64 \times 44$ & - & - & - & - & - & - & - & - & - & - & - & 77.4 & - \\
% & \textbf{MCAT} (Ours) & $64 \times 44$ & \textbf{78.5} & \textbf{90.5} & 91.6 & 86.9 & 84.4 & \textbf{80.8} & 83.3 & \textbf{87.5} & \textbf{88.0} & \textbf{85.0} & \textbf{73.0} & \textbf{84.5} & \textbf{5.5} \\\cline{2-16}
& \textbf{MCAT} (Ours) & $64 \times 44$ & 78.5 & 90.5 & 91.6 & 86.9 & 84.4 & 80.8 & 83.3 & 87.5 & 88.0 & 85.0 & 73.0 & 84.5 & 5.5 \\\cline{2-16}

& GLN \cite{hou2020gait} & $128 \times 88$ & 70.6 & 82.4 & 85.2 & 82.7 & 79.2 & 76.4 & 76.2 & 78.9 & 77.9 & 78.7 & 64.3 & 77.5 & 5.8 \\
& 3DLocal \cite{local3D} & $128 \times 88$ & 78.5 & 88.9 & 91.0 & 89.2 & 83.7 & 80.5 & 83.2 & 84.3 & 87.9 & 87.1 & 74.7 & 84.5 & 5.0 \\
% & \textbf{MCAT} (Ours) & $128 \times 88$ & \textbf{84.2} & \textbf{92.6} & \textbf{93.4} & \textbf{89.9} & \textbf{87.1} & \textbf{85.2} & \textbf{87.4} & \textbf{90.6} & \textbf{92.3} & \textbf{90.8} & \textbf{82.2} & \textbf{88.7} & \textbf{3.7} \\\hline
& STAR \cite{huang2022star} & $128 \times 88$ & 77.4 & 91.0 & 92.9 & 89.9 & 86.2 & 81.9 & 87.0 & \textbf{92.4} & \textbf{94.2} & 90.0 & 74.7 & 87.1 & 6.2 \\
& \textbf{MCAT} (Ours) & $128 \times 88$ & \textbf{84.2} & \textbf{92.6} & \textbf{93.4} & \textbf{89.9} & \textbf{87.1} & \textbf{85.2} & \textbf{87.4} & 90.6 & 92.3 & \textbf{90.8} & \textbf{82.2} & \textbf{88.7} & \textbf{3.7} \\\hline

\end{tabular}
\label{tab:casia-b}
\end{center}
\end{table*}

\begin{table*}[t]
\begin{center}
\footnotesize
\caption{Averaged rank-1 accuracies (\%) on OU-MVLP, excluding identical-view cases. Std denotes the performance sample standard deviation across 14 views. The results in the first 10 rows and the last 6 rows are conducted by keeping or removing invalid probe sequences that have no corresponding targets in the gallery set.}
\begin{tabular}{|c| c c c c c c c c c c c c c c |c|c|}
\hline
\multirow{2}{*}{Method} & \multicolumn{14}{c|}{Probe View} & \multirow{2}{*}{Mean} & \multirow{2}{*}{Std} \\\cline{2-15}
& $0^\circ$ & $15^\circ$ & $30^\circ$ & $45^\circ$ & $60^\circ$ & $75^\circ$ & $90^\circ$ & $180^\circ$ & $195^\circ$ & $210^\circ$ & $225^\circ$ & $240^\circ$ & $255^\circ$ & $270^\circ$ & &  \\\hline
GaitSet \cite{chao2021gaitset} & 81.3 & 88.6 & 90.2 & 90.7 & 88.6 & 89.1 & 88.3 & 83.1 & 87.7 & 89.4 & 89.7 & 87.8 & 88.3 & 86.9 & 87.9 & 2.6 \\
GaitPart \cite{fan2020gaitpart} & 82.6 & 88.9 & 90.8 & 91.0 & 89.7 & 89.9 & 89.5 & 85.2 & 88.1 & 90.0 & 90.1 & 89.0 & 89.1 & 88.2 & 88.7 & 2.3 \\
GLN \cite{hou2020gait} & 83.8 & 90.0 & 91.0 & 91.2 & 90.3 & 90.0 & 89.4 & 85.3 & 89.1 & 90.5 & 90.6 & 89.6 & 89.3 & 88.5 & 89.2 & 2.1 \\
GaitGL \cite{GaitGL} & 84.9 & 90.2 & 91.1 & 91.5 & 91.1 & 90.8 & 90.3 & 88.5 & 88.6 & 90.3 & 90.4 & 89.6 & 89.5 & 88.8 & 89.7 & 1.7 \\
3DLocal \cite{local3D} & 86.1 & 91.2 & \textbf{92.6} & \textbf{92.9} & \textbf{92.2} & 91.3 & 91.1 & 86.9 & 90.8 & \textbf{92.2} & \textbf{92.3} & \textbf{91.3} & \textbf{91.1} & 90.2 & 90.9 & 2.0 \\
GaitTransformer \cite{cui2022gaittransformer} & 87.9 & 91.3 & 91.6 & 91.7 & 91.6 & 91.3 & 91.1 & 90.3 & 90.4 & 90.8 & 91.0 & 90.6 & 90.3 & 90.0 & 90.7 & 1.0 \\
% Huang et al. \cite{huang2022enhanced} & 84.8 & 89.6 & 91.0 & 91.3 & 90.7 & 90.4 & 89.9 & 88.5 & 87.5 & 90.1 & 90.2 & 89.4 & 89.3 & 88.5 & 89.4 & 1.6 \\
Lagrange \cite{chai2022lagrange} & 85.9 & 90.6 & 91.3 & 91.5 & 91.2 & 91.0 & 90.6 & 88.9 & 89.2 & 90.5 & 90.6 & 89.9 & 89.8 & 89.2 & 90.0 & 1.4 \\
STAR \cite{huang2022star} & 85.5 & 90.0 & 91.4 & 91.6 & 90.5 & 90.7 & 90.2 & 88.0 & 88.5 & 90.5 & 90.7 & 89.7 & 89.7 & 88.9 & 89.7 & 1.5 \\
GaitBase \cite{fan2023opengait} & - & - & - & - & - & - & - & - & - & - & - & - & - & - & 90.8 & -\\
\textbf{MCAT} (Ours) & \textbf{88.5} & \textbf{91.5} & 91.7 & 92.0 & 91.8 & \textbf{91.4} & \textbf{91.2} & \textbf{90.3} & \textbf{90.9} & 91.1 & 91.2 & 90.8 & 90.7 & \textbf{90.4} & \textbf{91.0} & \textbf{0.9}  \\\hline \hline
GaitSet \cite{chao2021gaitset} & 84.5 & 93.3 & 96.7 & 96.6 & 93.5 & 95.3 & 94.2 & 87.0 & 92.5 & 96.0 & 96.0 & 93.0 & 94.3 & 92.7 & 93.3 & 3.5 \\
GaitPart \cite{fan2020gaitpart} & 88.0 & 94.7 & 97.7 & 97.6 & 95.5 & 96.6 & 96.2 & 90.6 & 94.2 & 97.2 & 97.1 & 95.1 & 96.0 & 95.0 & 95.1 & 2.7 \\
GLN \cite{hou2020gait} & 89.3 & 95.8 & 97.9 & 97.8 & 96.0 & 96.7 & 96.1 & 90.7 & 95.3 & 97.7 & 97.5 & 95.7 & 96.2 & 95.3 & 95.6 & 2.5 \\
GaitGL \cite{GaitGL} & 90.5 & 96.1 & 98.0 & 98.1 & 97.0 & 97.6 & 97.1 & 94.2 & 94.9 & 97.4 & 97.4 & 95.7 & 96.5 & 95.7 & 96.2 & 2.0 \\
3DLocal \cite{local3D} & - & - & - & - & - & - & - & - & - & - & - & - & - & - & 96.5 & - \\
\textbf{MCAT} (Ours) & \textbf{94.3} &\textbf{ 97.5} & \textbf{98.7} & \textbf{98.7} & \textbf{97.7} & \textbf{98.3} & \textbf{98.1} & \textbf{96.1} & \textbf{97.3} & \textbf{98.3} & \textbf{98.3} & \textbf{97.1} & \textbf{97.8} & \textbf{97.4} & \textbf{97.5} & \textbf{1.2} \\\hline
\end{tabular}
\label{tab:ou-mvlp}
\end{center}
\end{table*}

\begin{figure}[t]
    \centering
    \includegraphics[width=\linewidth]{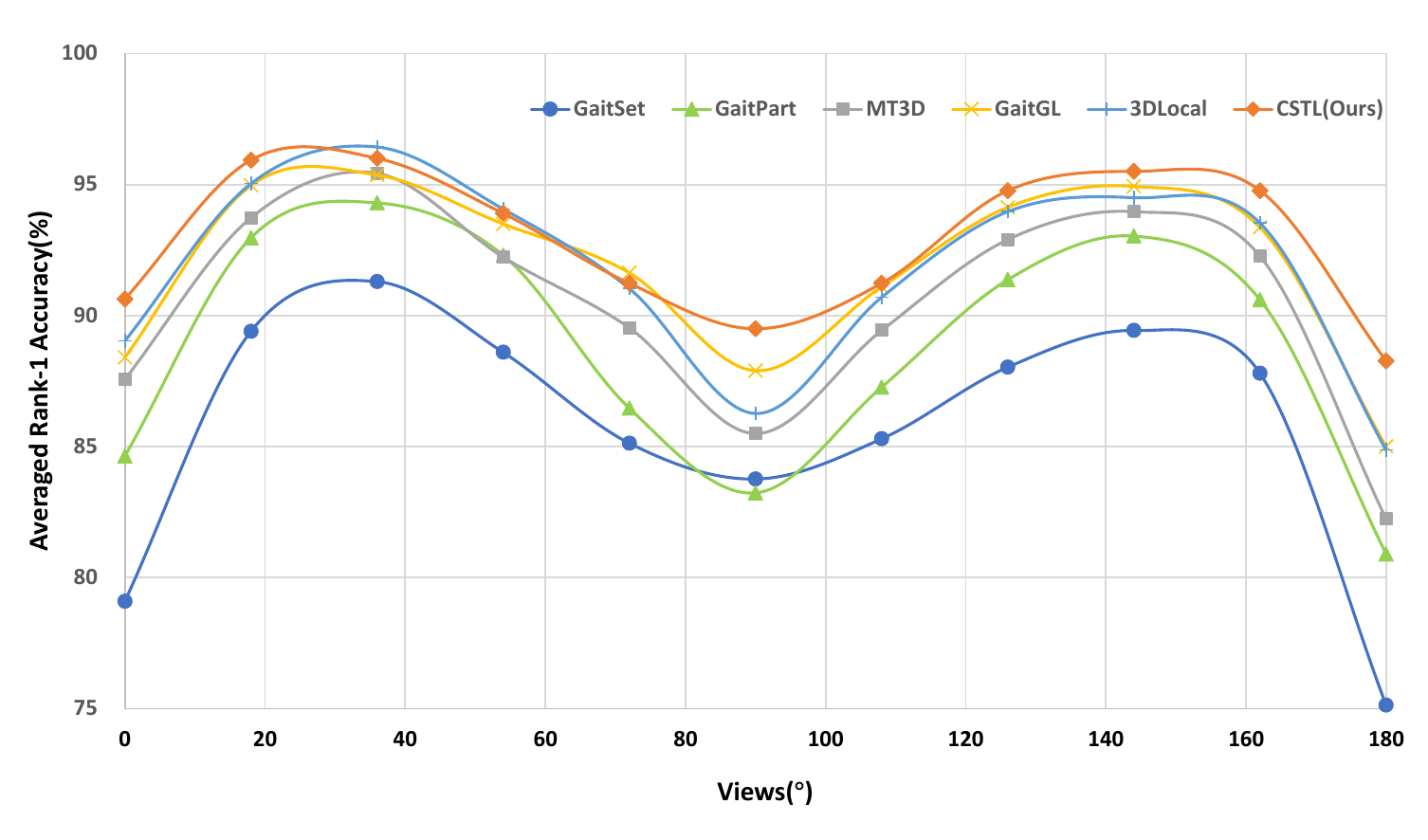}
    \caption{Multi-view performance comparison on CASIA-B using curve chart, in terms of averaged rank-1 accuracy.}
    \label{fig:casia-b-curve}
\end{figure}

\subsection{Comparison with the State-of-the-art Methods}

\noindent \textbf{CASIA-B.} Table. \ref{tab:casia-b} shows the comparison results between the proposed MCAT and current state-of-the-art methods in averaged rank-1 accuracies on CASIA-B dataset. Three walking conditions (NM, BG, CL) and 11 different camera views ($0^\circ-180^\circ$) are considered into performance evaluation. Three notable conclusions are summarized as: (1) MCAT outperforms all other methods in mean accuracy comparisons under all cases, which demonstrates the robustness and advantages. (2) It is natural that recognition performances would oscillate under different camera viewpoints. However, compared with other SOTA approaches, MCAT achieves almost the lowest performance standard deviation across 11 views under all walking conditions, which proves the robustness against viewpoint variations. (3) MCAT also shows robustness to resolution variations of gait sequences. Under different resolution settings of three walking conditions, MCAT achieves better mean recognition accuracies over current methods, which reveals that MCAT could adapt to different resolutions flexibly. Based on that, we use resolution setting of $64 \times 44$ in the rest of this paper since it achieves better tradeoff between performance and computation cost.

Further, we draw the performance curves of the 11 views of CASIA-B in Figure. \ref{fig:casia-b-curve}, which illustrates the accuracy fluctuations under cross-view scenarios. We find that: (1) All curves look like saddle-shape, which are roughly symmetric about the 90$^\circ$ view. This phenomenon reveals that sequences from symmetric views contain similar spatial-temporal information for recognition, which adheres to human intuitions. (2) Compared with the sequences from side view (90$^\circ$) and front (0$^\circ$) or back view (180$^\circ$), sequences from the squint views (36 $^\circ$ or 144 $^\circ$) achieve better performances, which may be attributed to the squint views incorporating rich visual clues from both side view and front or back view.

\begin{table}[t]
    \centering
    \caption{Averaged rank-1, rank-5, rank-10 and rank-20 accuracies (\%) on GREW, excluding identical-view cases.}
    \begin{tabular}{|c|c|c|c|c|}
        \hline
        Method & Rank-1 & Rank-5 & Rank-10 & Rank-20 \\\hline
        GEINet \cite{geinet} & 6.8 & 13.4 & 17.0 & 21.0 \\
        TS-CNN \cite{wu2016comprehensive} & 13.6 & 24.6 & 30.2 & 37.0 \\
        GaitSet \cite{chao2021gaitset} & 46.3 & 63.6 & 70.3 & 76.8 \\
        GaitPart \cite{fan2020gaitpart} & 44.0 & 60.7 & 67.3 & 73.5 \\
        \textbf{MCAT} (Ours) & \textbf{50.6} & \textbf{65.9} & \textbf{71.9} & \textbf{76.9} \\\hline
    \end{tabular}
    \label{tab:grew}
\end{table}

\noindent \textbf{OU-MVLP.} Table. \ref{tab:ou-mvlp} shows the comparison results between the proposed MCAT and current state-of-the-art methods in terms of averaged rank-1 accuracies on OU-MVLP. Our MCAT outperforms the existing methods under mean accuracy comparison, which proves the generalization capacity on a large-scale dataset. Particularly, though 3DLocal achieves competitive performance with MCAT under the setting of keeping invalid sequences, MCAT achieves much lower performance standard deviation than 3DLocal (0.9\% with 2.0\%), which demonstrates the stronger stability of viewpoint changes. Taking the performances under $0^\circ$ and $180^\circ$ as examples, MCAT marginally outperforms 3DLocal by $2.4\%$ and $3.4\%$ respectively. This phenomenon explains that MCAT has fewer preferences on certain camera viewpoints compared with current methods. Besides, by removing the invalid probe sequences, MCAT achieves much better accuracy compared with other methods.

\noindent \textbf{GREW.} Table. \ref{tab:grew} shows the comparison results between the proposed MCAT and current state-of-the-art methods. We can see sequences based method significantly outperforms Gait Energy Image (GEI) based method which indicates sequences based method has stronger ability to model the gait pattern in real scenarios. Compared to other temporal modeling methods, MCAT, which has a richer representation of temporal features, has a large margin in all of the test settings. MCAT achieves the best performances under all comparison settings, which proves the gait modeling capacity of MCAT under complex real-world scenarios.

\subsection{Ablation Study}
In order to evaluate the exact effectiveness of MCAT, ablation experiments are conducted on CASIA-B to study the proposed components.

\begin{table}[ht]
    \footnotesize
    \centering
    \caption{Study of the effectiveness of modules in MCAT on CASIA-B in terms of averaged rank-1 accuracy (\%). For the sake of simplicity, we use MSTE to denote multi-scale temporal extraction.}
    \begin{tabular}{|c|c|c|c|c|}
        \hline
        \multirow{2}{*}{Model} & \multicolumn{4}{c|}{Rank-1 Accuracy} \\\cline{2-5}
        & NM & BG & CL & Mean\\\hline
        GaitSet \cite{chao2021gaitset} & 96.1 & 90.8 & 70.3 & 85.7 \\\hline
        GaitPart \cite{fan2020gaitpart} & 96.2 & 91.5 & 78.7 & 88.8 \\\hline
        \multicolumn{5}{|c|}{\textbf{Ours}} \\\hline
        Baseline & 95.3 & 88.7 & 72.1 & 85.4 \\\hline
        Baseline + MSTE & 96.6 & 91.1 & 81.0 & 89.6 \\\hline
        Baseline + MSTE + ATA & 98.3 & 94.0 & 81.5 & 91.3 \\\hline
        Baseline + MSTE + SSFL & 97.7 & 93.0 & 83.8 & 91.5 \\\hline
        MCAT & \textbf{98.5} & \textbf{94.8} & \textbf{84.5} & \textbf{92.6} \\\hline
    \end{tabular}
    \label{tab:spatio-temporal comparison}
\end{table}

\noindent \textbf{Impact of Spatio-Temporal Modeling.} The individual effects of spatial and temporal modeling are presented in Table. \ref{tab:spatio-temporal comparison}. The baseline refers to the 4-layer CNN with a feature division, while using a BA+ loss for supervision. Several noteworthy observations can be summarized as: (1) Compared to spatial modeling network, i.e. GaitSet \cite{chao2021gaitset}, our baseline achieves similar mean accuracy under three conditions (85.7\% and 85.4\%), which proves the competitive spatial learning capacity of them. However, with the utilization of MSTE, our method achieves significant accuracy improvement over GaitSet \cite{chao2021gaitset} (+3.9\%), which indicates the superiority of modeling multi-scale temporal context based on spatial feature extraction. (2) Compared with the multi-scale temporal modeling method, i.e., GaitPart, our method achieves higher performances when applying MSTE with ATA (+2.5\%) and MSTE with SSFL (+2.7\%), which verifies the effectiveness of adaptive temporal aggregation and salient spatial mining based on multi-scale temporal clues. (3) Applying both spatial and temporal modeling achieves the best results, which verifies the complementary properties provided by SSFL and ATA modules.

\noindent \textbf{Effectiveness of Local and Global Relation Modeling.} Table. \ref{tab:relation modeling} investigates the effectiveness of the proposed local and global relation modeling strategies. In order to investigate the superiority of the proposed local relation modeling module, we use max pooling and attention \cite{cstl} method for comparsion. We find that, using an FC achieves the best performance over other methods, which proves its superiority for fusing multi-scale features locally. Therefore, we take an FC as local relation modeling in the final version. Further, local-with-global joint modeling achieves the best performance, which demonstrates the complementary properties introduced by local and global relation modelings.

\begin{table}[ht]
    \centering
    \scriptsize
    \setlength \tabcolsep{4pt}
    \caption{Study the effectiveness of local and global relation modeling on CASIA-B in terms of averaged rank-1 accuracy (\%).}
    \begin{tabular}{|c|c|c|c|c|c|c|c|}
        \hline
        \multicolumn{3}{|c|}{Local Relation} & \multirow{3}{*}{Global Relation} & \multicolumn{4}{c|}{Rank-1 Accuracy} \\\cline{1-3} \cline{5-8}
        Max & \multirow{2}{*}{FC} & \multirow{2}{*}{Attention} & & \multirow{2}{*}{NM} & \multirow{2}{*}{BG} & \multirow{2}{*}{CL} & \multirow{2}{*}{Mean} \\
        Pooling & & & & & & & \\\hline
        \checkmark & & & & 97.9 & 93.8 & 84.4 & 92.0 \\\hline
        & \checkmark & & & 98.1 & 93.9 & 84.5 & 92.2 \\\hline
        & & \checkmark & & 97.9 & 93.8 & 82.8 & 91.5 \\\hline
        & & & \checkmark & 98.3 & 94.0 & 81.8 & 91.4 \\\hline
        & \checkmark & & \checkmark & \textbf{98.5} & \textbf{94.8} & \textbf{84.5} & \textbf{92.6} \\\hline
    \end{tabular}
    \label{tab:relation modeling}
\end{table}

\noindent \textbf{Comparison of Spatial Selection
Strategies} In order to investigate the effectiveness of our spatial learning module for supplementing corrupted spatial features, we conduct two more experiments for comparison: (1) We replace SSFL with a random frame selection to demonstrate that not each frame has good spatial features. (2) We set the number of selected parts as 1 in SSFL. In this situation, SSFL turns to be a frame-level feature selection instead of part-level feature selection.

\begin{table}[ht]
    \centering
    \caption{Comparisons of spatial selection strategies on CASIA-B in terms of averaged rank-1 accuracy (\%).}
    \begin{tabular}{|c|c|c|c|c|}
    \hline
    \multirow{2}{*}{Methods} & \multicolumn{4}{c|}{Rank-1 Accuracy} \\\cline{2-5}
    & NM & BG & CL & Mean \\\hline
    random frame & 98.1 & 94.0 & 82.9 & 91.7 \\\hline
    SSFL (frame-level) & 98.5 & 94.5 & 83.7 & 92.2 \\\hline
    SSFL & \textbf{98.5} & \textbf{94.8} & \textbf{84.5} & \textbf{92.6} \\\hline
    \end{tabular}
    \label{tab: spatial selection}
\end{table}

As shown in Table. \ref{tab: spatial selection}, we notice that: SSFL outperforms the other two strategies, which proves the spatial learning capability of our method. On one hand, random frame selection is probably incapable of obtaining high quality spatial features due to the randomness. On the other hand, although frame-level spatial selection achieves better performance than random frame selection, it still limits the diverse discriminative expression of local parts, especially considering the occlusion of motion and change of camera viewpoints. Compared to the above strategies, our SSFL extracts spatial clues in a fine-grained manner and utilizes the inherent motion characteristics to leverage rich visual clues across the sequence.

\begin{table}[ht]
    \centering
    \caption{Study on the impacts of importance evaluation methods and number of selected spatial parts in SSFL on CASIA-B in terms of rank-1 averaged accuracy (\%).}
    \begin{tabular}{|c|c|c|c|c|c|}
        \hline
        Method & Selected Groups & \multicolumn{4}{c|}{Rank-1 Accuracy}  \\\hline
        MLP & 1 & 98.1 & 93.6 & 83.5 & 91.7 \\\hline
        \multirow{4}{*}{MHSA} & 1 & 98.3 & 93.7 & 83.6 & 91.9 \\\cline{2-6}
        & 2 & 98.3 & 93.8 & 83.5 & 91.8 \\\cline{2-6}
        & 4 & \textbf{98.5} & \textbf{94.8} & \textbf{84.5} & \textbf{92.6} \\\cline{2-6}
        & 8 & 98.5 & 94.2 & 83.6 & 92.1 \\\hline
    \end{tabular}
    \label{tab:importance evaluation}
\end{table}

\noindent \textbf{Investigation on the impact factors of SSFL.} Table \ref{tab:importance evaluation} investigates the impact factors in SSFL on CASIA-B dataset. Especially, the first experiment is conducted by using a MLP to produce the part scores of each frame locally, which can only select a group of salient parts. We can notice that: (1) MHSA outperforms MLP when selecting only one group of spatial parts, which illustrates the superiority of constructing the importance map globally. (2) When selecting more groups of spatial parts by MHSA, the recognition performance first improves continually, then drops. This phenomenon explains that the number of high quality salient parts in each sequence is limited, thus we may obtain low-quality parts by selecting redundant groups, which hurts the recognition performance. To achieve the best performance, we select 4 groups in our final version.

\begin{figure}[ht]
    \centering
    \includegraphics[width=\linewidth]{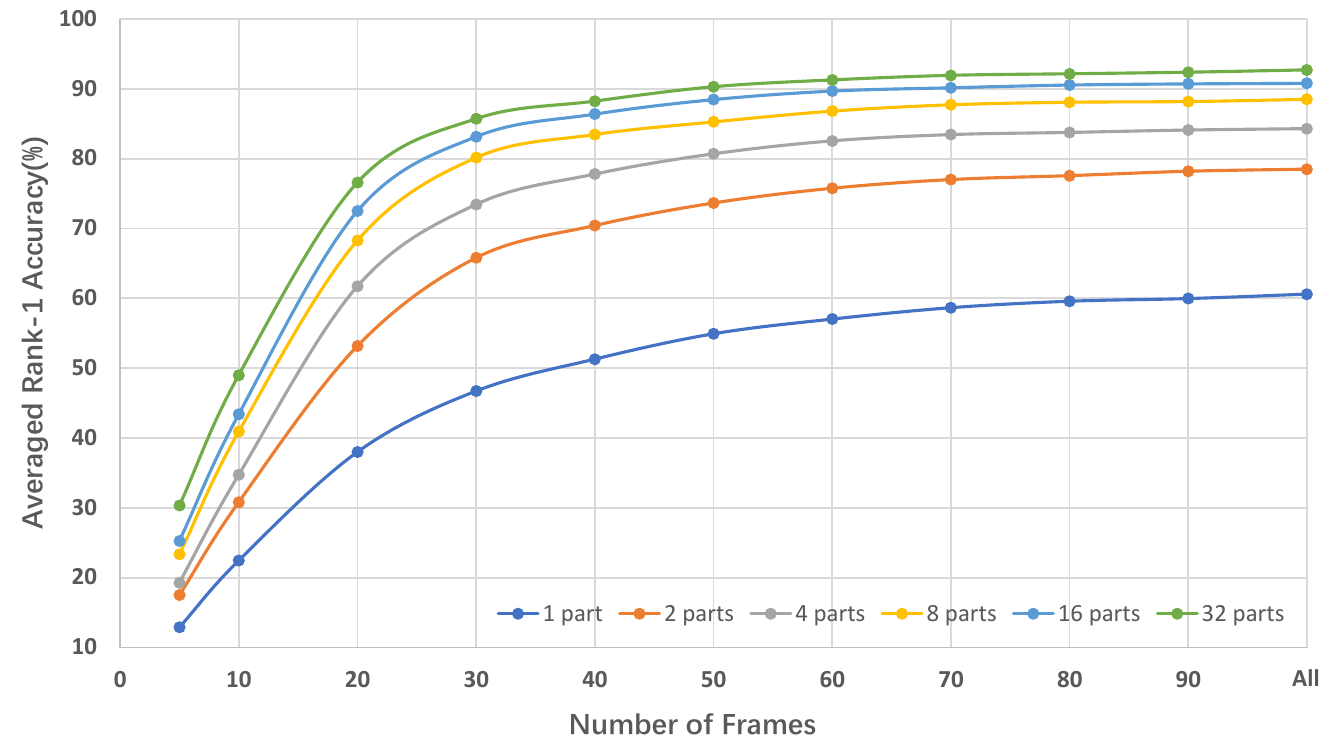}
    \caption{Study on the impact of different part division numbers and frame numbers on CASIA-B \cite{yu2006framework} in terms of averaged rank-1 accuracy under NM, BG and CL conditions.}
    \label{fig:part_analysis}
\end{figure}

\noindent \textbf{Ablation Study on Part Division Numbers and Frame Numbers.} In order to investigate the effects of the part division number $K$ and frame number $N$, we conduct ablation experiments with different number of $K$ and $N$. Particularly, 32 is the largest number for $K$, since the output feature dimension is $32 \times 22$. And the maximum $N$ is set as the number of all frames in each sequence. As shown in Figure. \ref{fig:part_analysis}, we can see that the accuracy improves continually with the increasing of number of parts and number of frames, which indicates that: (1) More fine-grained part division provides richer clues for modeling spatial local features, which further satisfies the diverse motion expression of different body parts. (2) More frames contain more abundant information for constructing temporal contextual communications, which enables the network to extract more discriminative temporal clues and mine more salient spatial parts.

Therefore, to achieve the best performance, we set the $K$ as 32, and use all frames during test stage.

\subsection{Practical Scenarios}
In this section, we consider two new experimental settings, which are closer to real-world applications.  (1) Because of the possible insufficiency of training data, models may be tested under unseen views in real life. (2) People may walk in arbitrary directions anytime and anywhere, thus one sequence may be composed of frames from different views.

\begin {figure}[t]
    \centering
    \includegraphics[width=\linewidth]{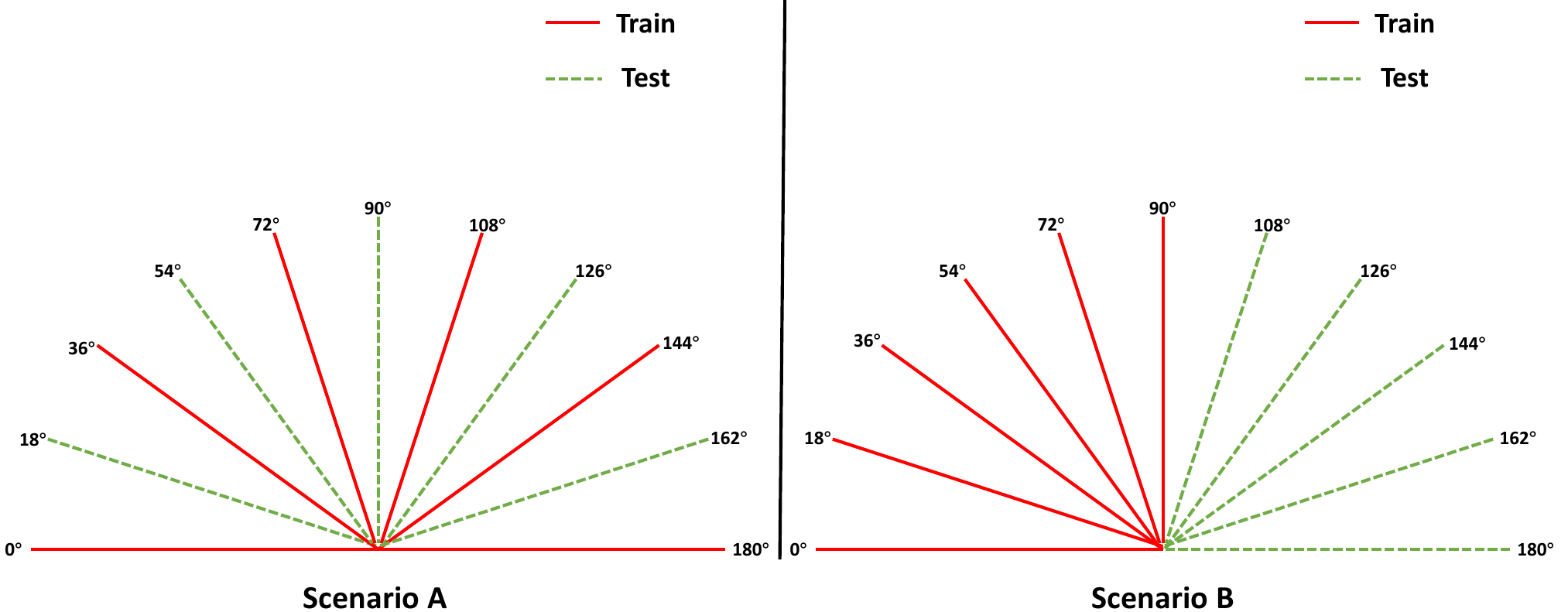}
    \caption{Illustration of testing models under unseen views.}
    \label{fig:unseen scenarios}
\end{figure}

\noindent \textbf{Testing Under Unseen Views.} As shown in Figure. \ref{fig:unseen scenarios}, we consider two possible scenarios for testing the view generalization capacity of our method. Specially, scenario A corresponds to the case that the training dataset covers the view range in the test dataset, but does not include some certain views. In contrast, scenario B refers to the case that views in the training dataset and test dataset are in different ranges. Intuitively, scenario B is harder than scenario A.

As reported in Table. \ref{tab:unseen_view}, the accuracy decreases under unseen views as expected. However, compared with the baseline, our method achieves stronger robustness against unseen scenarios, whose performances degrade 2.5\% under scenario A and 8.1\% under scenario B. These results demonstrate the spatial-temporal modeling and view generalization capacities of the proposed modules.

\begin{figure*}[t]
    \centering
    \includegraphics[width=0.95\linewidth]{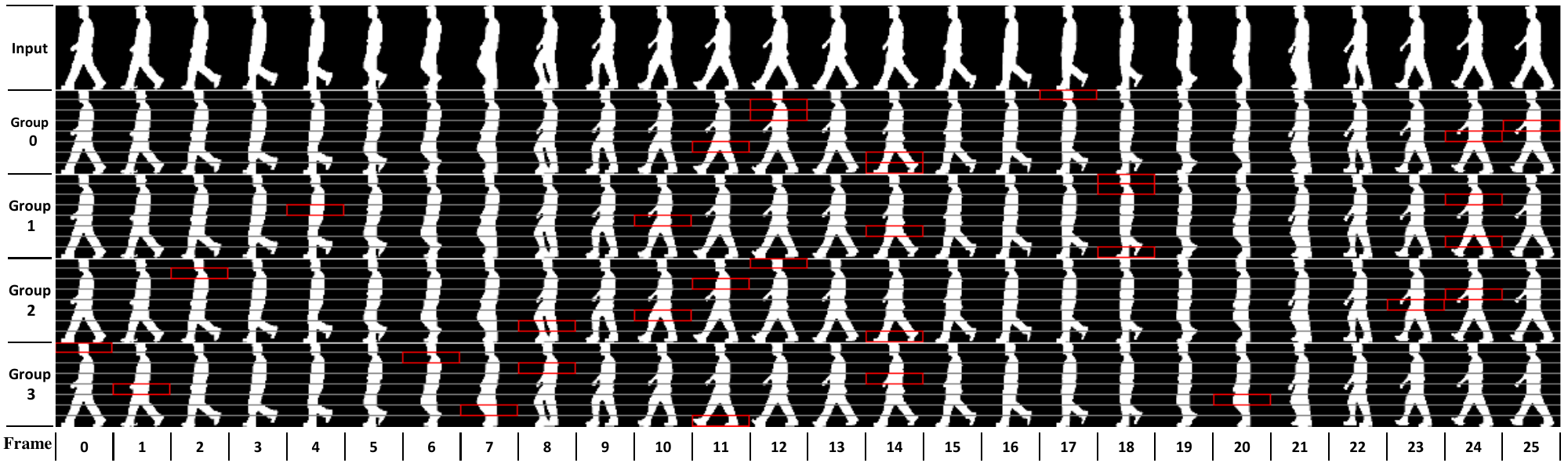}
    \caption{Illustration of spatial salient feature learning. We select four groups of salient local parts. The red boxes indicate selected parts.}
    \label{fig:part_selection}
\end{figure*}

\begin{table}[ht]
    \centering
    \caption{Performance comparison under unseen views on CASIA-B in terms of averaged rank-1 accuracy (\%) under all walking conditions, excluding identical-view cases. Particularly, the default setting denotes that test views are in accordance with train views.}
    \begin{tabular}{|c|c|c|c|}
    \hline
    % \multirow{2}{*}{Method} & \multirow{2}{*}{Scenario A} & \multirow{2}{*}{Scenario B} & Default \\
    % & & & Setting\\\hline
    Method & Default Setting & Scenario A & Scenario B \\\hline
    Baseline & 85.4 & 81.9 & 76.3 \\\hline 
    Ours & 92.6 & 90.1 & 84.5 \\\hline
    \end{tabular}
    \label{tab:unseen_view}
\end{table}

\noindent \textbf{Sequences include different views} As shown in Table. \ref{tab:multi_view}, we conduct 5 more experiments that combine frames from different views into one sequence. Particularly, each sequence is composed of frames from a pair of views, which are defined by the view difference. Taking the view difference of $90^\circ$ as an example, the corresponding pairs are: $0^\circ$ \& $90^\circ$, $18^\circ$ \& $108^\circ$, $36^\circ$ \& $126^\circ$, $54^\circ$ \& $144^\circ$, $72^\circ$ \& $162^\circ$, $90^\circ$ \& $180^\circ$. And for eliminating the effects of sequence length as far as possible, we only sample half of each sequence from each view. Particularly, the sequences in the probe set are composed of frames from different views, and the sequences in the gallery set are unchanged.

\begin{table}[ht]
    \centering
    \footnotesize
    \caption{Performance comparison using sequences include different views on CASIA-B in terms of averaged rank-1 accuracy (\%) under all walking conditions, excluding identical-view cases.}
    \begin{tabular}{|c|c|c|c|c|c|c|}
        \hline
        View Difference & $18^\circ$ & $36^\circ$ & $54^\circ$ & $72^\circ$ & $90^\circ$ & Single View  \\\hline
        Baseline & 88.1 & 89.6 & 89.8 & 90.2 & 90.2 & 85.4 \\\hline
        Ours & 93.0 & 94.6 & 95.3 & 95.7 & 95.9 & 92.6 \\\hline
    \end{tabular}
    \label{tab:multi_view}
\end{table}

As given Table. \ref{tab:multi_view}, our method marginally outperforms the baseline with all view pairs, which further proves the effectiveness of our method under novel scenarios. Interestingly, we find that: (1) Using frames from different views achieves higher performances than using frames from one single view, which reveals that richer clues could be obtained in different views. (2) Using frames from larger view differences could consistently achieve higher performances, which reflects that more complementary information is provided in larger view-difference pairs. 

Consequently, we argue that combine frames from different views could facilitate gait recognition in real-world scenarios effectively.

% \subsection{Visualization}
% \begin{figure*}[t]
%     \centering
%     \includegraphics[width=0.9\linewidth]{fig/heatmap.pdf}
%     \caption{Illustration of attention heatmaps from the last layer in the backbone. The color bar on the right indicates the attention distribution for different colors. Best viewed in color.}
%     \label{fig:heatmap}
% \end{figure*}
\noindent \textbf{Salient Spatial Part Selection.} In order to better understand the positive effects of SSFL, we give an example in Figure. \ref{fig:part_selection}, where we set the number of selected parts as 8 for better visualization. We can notice that: (1) Directly perceived through the senses, SSFL tends to select parts with clear representation and complete appearance features, which are not affected by body overlaps and clothing occlusions, e.g., from frame 8 to frame 14 and from frame 23 to frame 25. In contrast, frames with body overlaps are less selected, e.g., from frame 3 to frame 7 and from frame 19 to frame 21. (2) Various groups select a particular body part from different frames, showcasing the diverse focus in several aspects of MHSA. For example, the head in group 1 is selected from frame 18 while in group 3 is selected from frame 0.

In this way, we can obtain high quality spatial features, which both remedies the negative influences caused by temporal operations and enhances the robustness of our network against occlusion variations.

% \noindent \textbf{Attention Visualization.} Figure. \ref{fig:heatmap} illustrates the attention maps from the last layer in the backbone of the baseline and our method. Intuitively, the baseline network mainly pays attention on the most discriminative parts of the human body, i.e., the head and the legs, whose focus is concentrated on certain regions. In contrast, the focus of our method is relatively scattered, which not only notices the parts that the baseline focus on, e.g., the legs, but also attends on the torso of the body that contains supplementary clues to recognize subjects. Therefore, compared with the baseline, our method could extract richer identity-related information thus achieves higher performance.

\begin{figure}[ht]
\centering
\subfigure[Baseline]{
\begin{minipage}[t]{0.45\linewidth}
\label{baseline_tsne}
\centering
\includegraphics[width=1.8in]{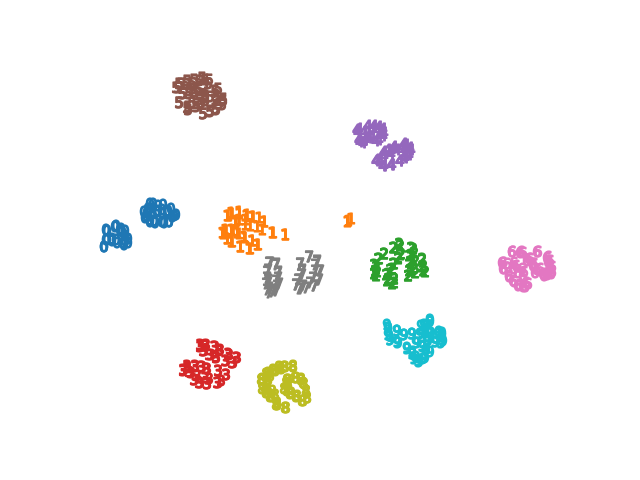}
\end{minipage}%
}%
% \rulesep
\subfigure[Ours]{
\begin{minipage}[t]{0.4\linewidth}
\label{ours_tsne}
\centering
\includegraphics[width=1.8in]{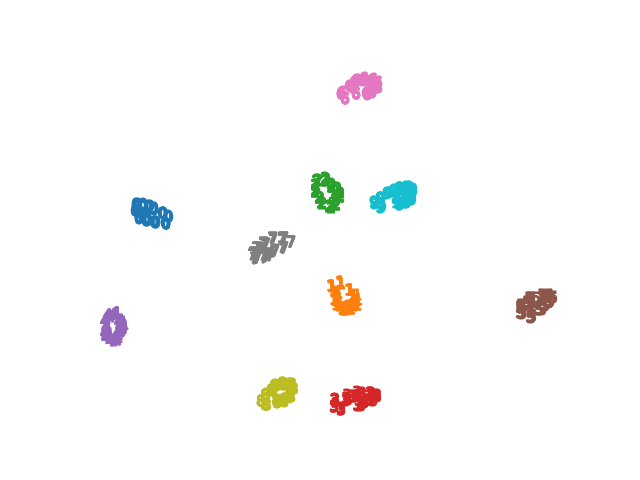}
\end{minipage}%
}%
\centering
\caption{tSNE visualization examples of the baseline and our proposed model on CASIA-B test dataset. Different numbers with different colors indicate different identities. Best viewed with zooming in.}
\end{figure}

\noindent \textbf{Feature Distribution.} We choose ten identities from CASIA-B test dataset to visualize feature distributions by t-SNE \cite{van2008visualizing}. Comparing the feature distributions of baseline and our method, we notice that, in Figure. \ref{baseline_tsne}, the feature distributions of different subjects are closer to each other thus identities are harder to distinguish. Differently, in Figure. \ref{ours_tsne}, the feature distributions of different subjects are more scattered to each other thus identities are more distinguishable, which proves the discriminant feature learning ability of our method.

\section{Conclusion}
\label{conclusion}
In this paper, we propose a multi-scale context-aware network with transformer (MCAT) for gait recognition. 
% context-sensitive temporal feature learning (CSTL) network for gait recognition
MCAT extracts temporal features with multiple scales and captures salient spatial clues for achieving strong spatio-temporal modeling ability. Specifically, diverse temporal features in three scales are introduced in MCAT, and local-to-global temporal relations are considered based on these temporal information for adaptive temporal aggregation. Besides, discriminative spatial parts are selected across the sequence to remedy the spatial feature corruption. Extensive experiments on three public datasets verify the superiority and the real-world application potentials of our method. In addition, we argue that the insights of learning spatial and temporal features in a supplementary manner could be also applied in other human action-related tasks, e.g., video-based person re-identification. We leave this for future work.  

{\small
\bibliographystyle{IEEEtran}
\bibliography{main}
}

\begin{IEEEbiography}[{\includegraphics[width=1in,height=1.25in,clip,keepaspectratio]{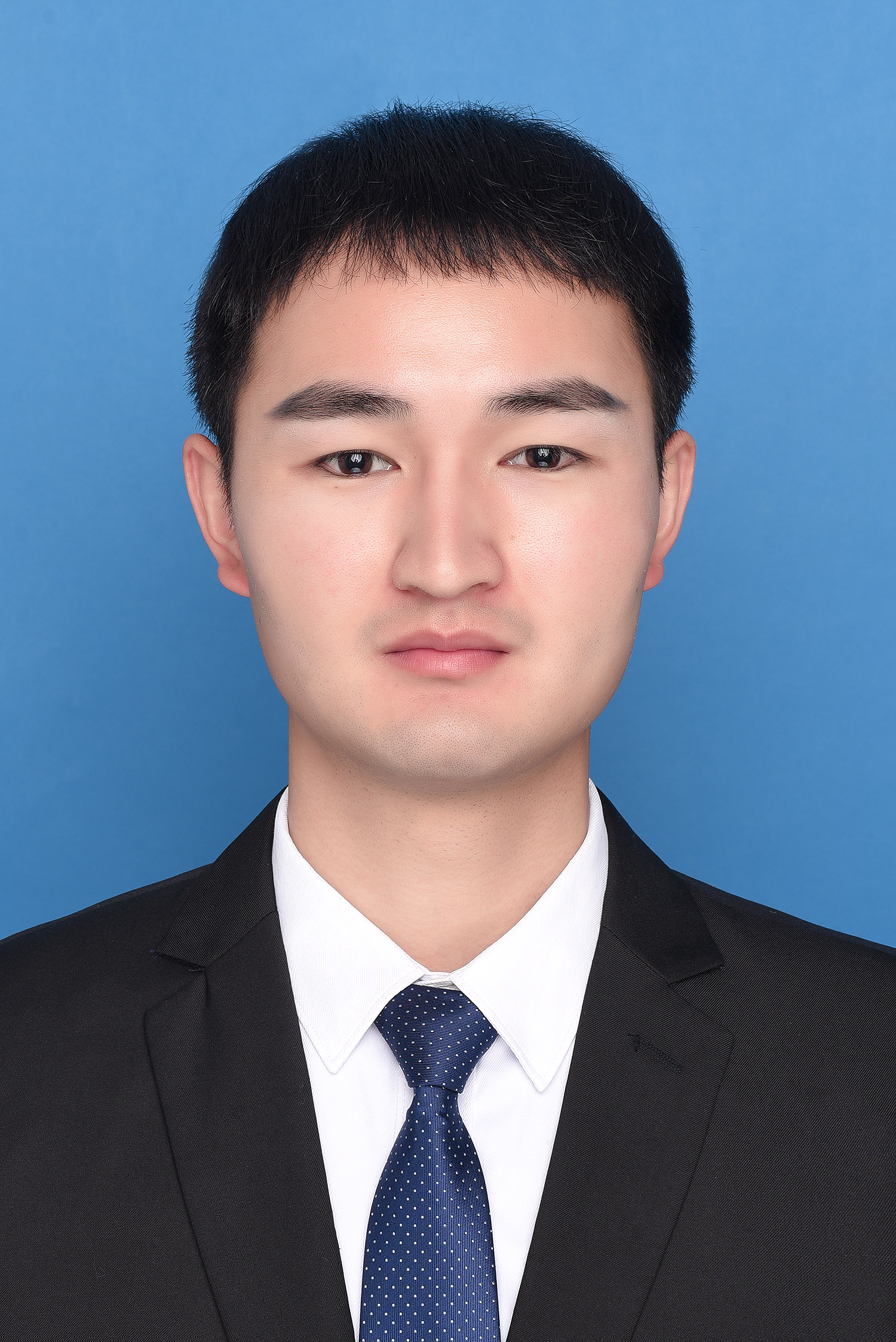}}]{Duowang Zhu} received the B.S. degree in School of Electronic Information and Communications from Huazhong University of Science and Technology (HUST), Wuhan, China, in 2020. Now, he is pursuing the M.S. degree in School of Electronic Information and Communications from Huazhong University of Science and Technology (HUST), Wuhan, China. His current research areas include computer vision and machine learning.
\end{IEEEbiography}

\begin{IEEEbiography}[{\includegraphics[width=1in,height=1.25in,clip,keepaspectratio]{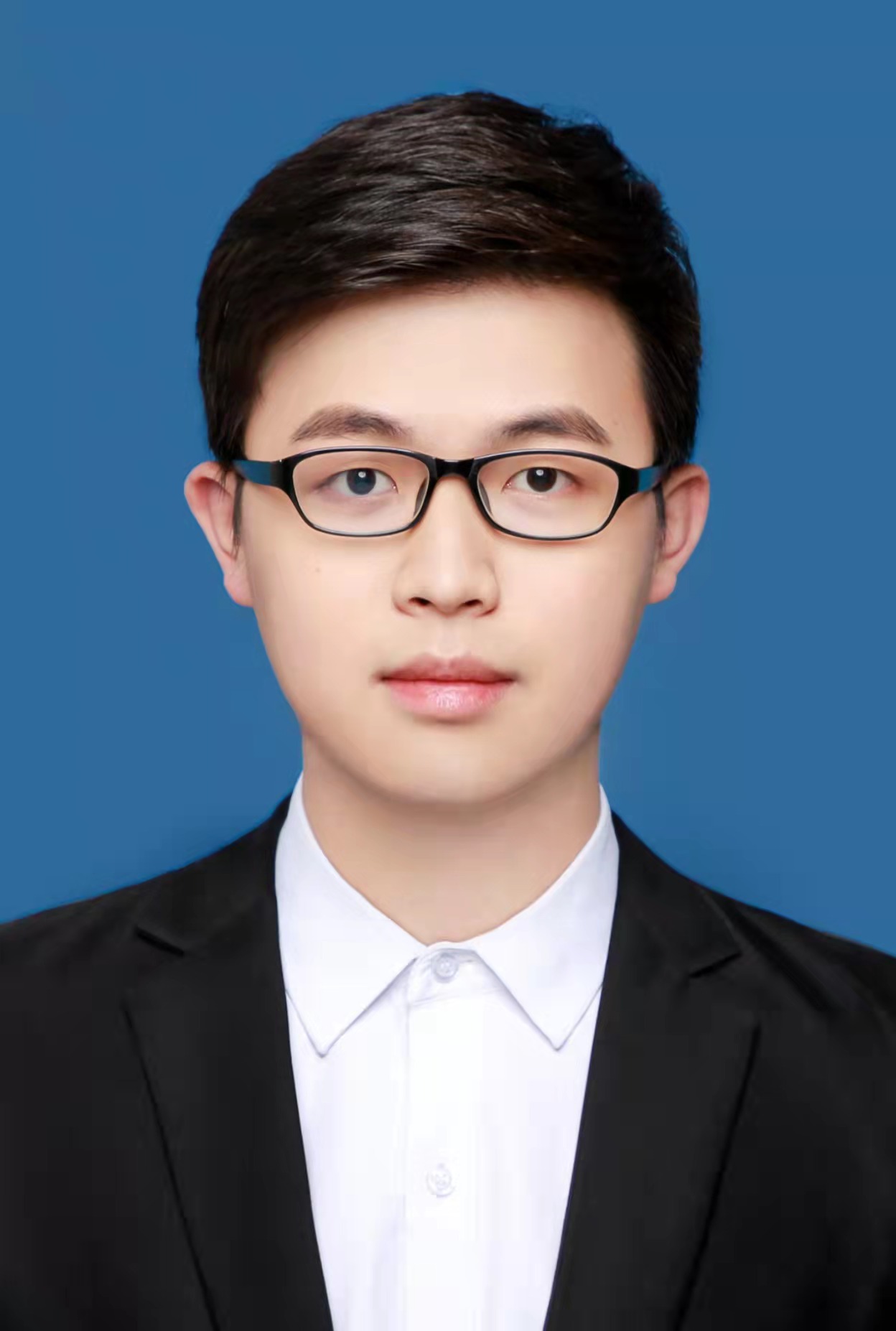}}] {Xiaohu Huang} received the B.S. degree in School of Electronic Information and Communications from Huazhong University of Science and Technology (HUST), Wuhan, China, in 2020. Now, he is pursuing the M.S. degree in School of Electronic Information and Communications from Huazhong University of Science and Technology (HUST), Wuhan, China. His current research areas include computer vision and machine learning.
\end{IEEEbiography}

\begin{IEEEbiography}[{\includegraphics[width=1in,height=1.25in,clip,keepaspectratio]{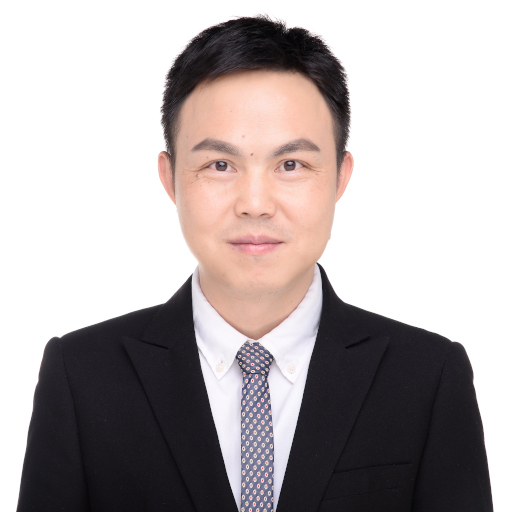}}] {Xinggang Wang (M’17)} received the B.S. and Ph.D. degrees in Electronics and Information Engineering from Huazhong University of Science and Technology (HUST), Wuhan, China, in 2009 and 2014, respectively. He is currently an Associate Professor with the School of Electronic Information and Communications, HUST. His research interests include computer vision and machine learning. He services as associate editors for Pattern Recognition and Image and Vision Computing journals and an editorial board member of Electronics journal.
\end{IEEEbiography}

\begin{IEEEbiography}[{\includegraphics[width=1in,height=1.25in,clip,keepaspectratio]{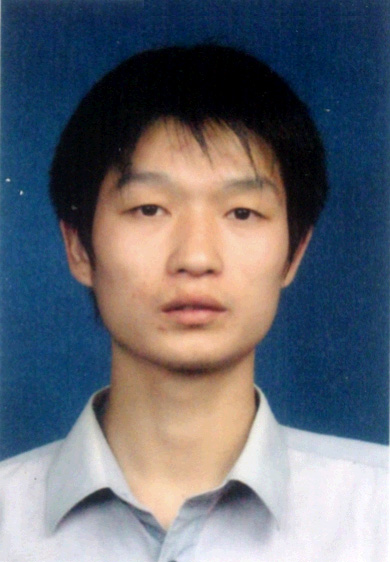}}]{Bo Yang} received the Master degree in School of mathematics and statistics form Wuhan University,Wuhan, China. He is currently the senior engineer of Wuhan FiberHome Digital Technology Co., Ltd. His research interests include computer vision and data mining.
\end{IEEEbiography}

\begin{IEEEbiography}[{\includegraphics[width=1in,height=1.25in,clip,keepaspectratio]{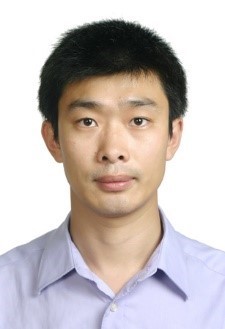}}]{Botao He} received the Ph.D. degree in School of Optical and Electronic Information from Huazhong University of Science and Technology (HUST), Wuhan, China. He is currently the deputy general manager of CICT Mobile Communication Technology Company Ltd.. His research interests include computer vision and data mining.
\end{IEEEbiography}

\begin{IEEEbiography}[{\includegraphics[width=1in,height=1.25in,clip,keepaspectratio]{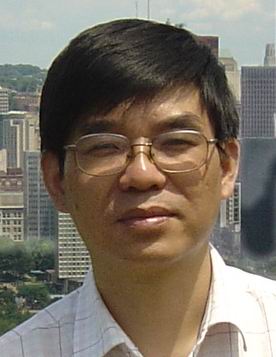}}] {Wenyu Liu (SM'15)} received the B.S. degree in Computer Science from Tsinghua University, Beijing, China, in 1986, and the M.S. and Ph.D. degrees, both in Electronics and Information Engineering, from Huazhong University of Science and Technology (HUST), Wuhan, China, in 1991 and 2001, respectively. He is now a professor and associate dean of the School of Electronic Information and Communications, HUST. His current research areas include computer vision, multimedia, and machine learning.
\end{IEEEbiography}

\begin{IEEEbiography}[{\includegraphics[width=1in,height=1.25in,clip,keepaspectratio]{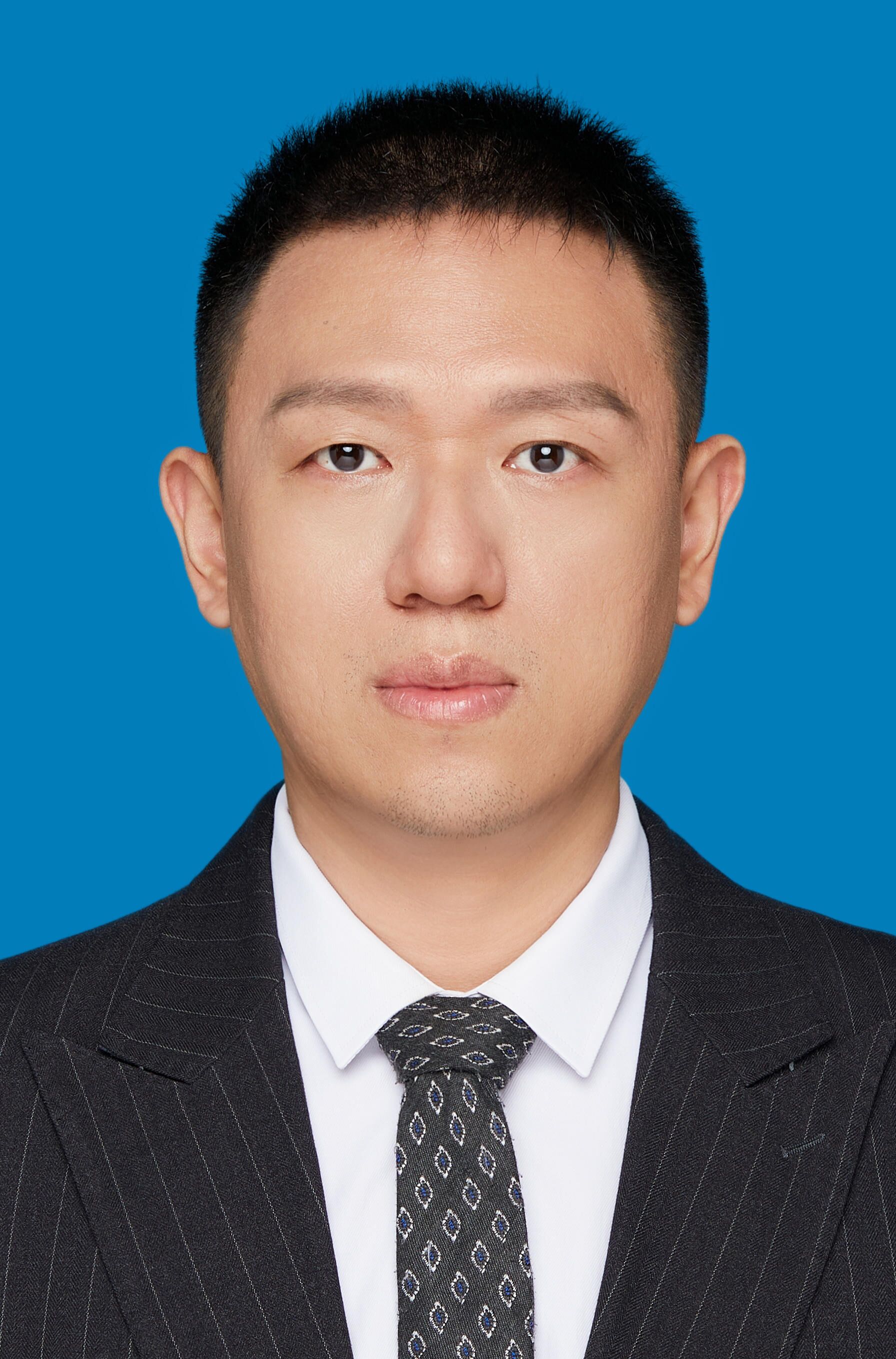}}] {Bin Feng} received the B.S. and Ph.D. degrees in School of Electronics and Information Engineering from Huazhong University of Science and Technology (HUST), Wuhan, China, in 2001 and 2006, respectively. He is currently an Associate Professor with the School of Electronic Information and Communications, HUST. His research interests include computer vision and intelligent video analysis.
\end{IEEEbiography}

\vfill

\end{document}